\documentclass[nohyperref]{article}

\usepackage{microtype}
\usepackage{graphicx}
\usepackage{subfigure}
\usepackage{booktabs} 
\usepackage[
  bookmarks=false,
  pdfpagelabels=false,
  hyperfootnotes=false,
  hyperindex=false,
  pageanchor=false, hidelinks, 
]{hyperref}

\makeatletter
\let\saved@hyper@linkurl\hyper@linkurl
\let\saved@hyper@link@\hyper@link@
\AtBeginDocument{%
  \NoHyper
  \let\hyper@linkurl\saved@hyper@linkurl 
  \let\hyper@link@\saved@hyper@link@ 
}
\makeatother

\usepackage{algorithm}
\usepackage{algorithmic}
\usepackage{dsfont}     
\usepackage{xcolor}
\usepackage{tcolorbox}

\usepackage{natbib}
\setcitestyle{authoryear,round,citesep={,},aysep={},yysep={;}}

\usepackage[accepted,nohyperref]{icml2023}

\usepackage{amsmath}
\usepackage{amssymb}
\usepackage{mathtools}
\usepackage{amsthm}
\DeclareMathOperator{\E}{\mathbb{E}}

\newcommand{\Lim}[1]{\raisebox{0.5ex}{\scalebox{0.98}{$\displaystyle \lim_{#1}\;$}}}

\usepackage[capitalize,noabbrev]{cleveref}

\definecolor{darkblue}{rgb}{0, 0, 0.55}
\setcitestyle{aysep={},yysep={,},citesep={,}}
\theoremstyle{plain}
\newtheorem{theorem}{Theorem}[section]
\newtheorem{proposition}[theorem]{Proposition}
\newtheorem{lemma}[theorem]{Lemma}
\newtheorem{corollary}[theorem]{Corollary}
\theoremstyle{definition}

\theoremstyle{remark}

\usepackage[textsize=tiny]{todonotes}

\icmltitlerunning{Correcting discount-factor mismatch in on-policy policy gradient methods}

\begin{document}

\twocolumn[
\icmltitle{Correcting discount-factor mismatch in on-policy policy gradient methods}



\icmlsetsymbol{equal}{*}

\begin{icmlauthorlist}
\icmlauthor{Fengdi Che}{yyy}
\icmlauthor{Gautham Vasan}{yyy}
\icmlauthor{A. Rupam Mahmood}{yyy,yyyy}
\end{icmlauthorlist}

\icmlaffiliation{yyy}{Department of Computing Science, University of Alberta, Edmonton, Canada}
\icmlaffiliation{yyyy}{CIFAR AI Chair, Amii,Department of Computing Science, University of Alberta, Edmonton, Canada}

\icmlcorrespondingauthor{Fengdi Che}{fengdi@ualberta.ca}

\icmlkeywords{On-policy policy gradient methods, discount factor mismatch, reinforcement learning}

\vskip 0.3in
]



\printAffiliationsAndNotice{}  

\begin{abstract}
The policy gradient theorem gives a convenient form of the policy gradient in terms of three factors: an action value, a gradient of the action likelihood, and a state distribution involving discounting called the \emph{discounted stationary distribution}. But commonly used on-policy methods based on the policy gradient theorem ignores the discount factor in the state distribution, which is technically incorrect and may even cause degenerate learning behavior in some environments. 
An existing solution corrects this discrepancy by using $\gamma^t$ as a factor in the gradient estimate. However, this solution is not widely adopted and does not work well in tasks where the later states are similar to earlier states. 
We introduce a novel distribution correction to account for the discounted stationary distribution that can be plugged into many existing gradient estimators. 
Our correction circumvents the performance degradation associated with the $\gamma^t$ correction with a lower variance.
Importantly, compared to the uncorrected estimators, our algorithm provides improved state emphasis to evade suboptimal policies in certain environments and consistently matches or exceeds the original performance on several OpenAI gym and DeepMind suite benchmarks.$^\dagger$
\let\thefootnote\relax\footnote{$^\dagger$ Source code: \href{https://github.com/FengdiC/Averaged_PPO.git}{\textcolor{darkblue}{Averaged PPO}} and \href{https://github.com/FengdiC/Averaged_Dist_Corr.git}{\textcolor{darkblue}{the rest}}.} 
\end{abstract}

\begin{figure}[ht]
    \centering
    \vspace*{-2mm}
    \hspace*{-2mm}
    \includegraphics[width=0.35\textwidth
    ]{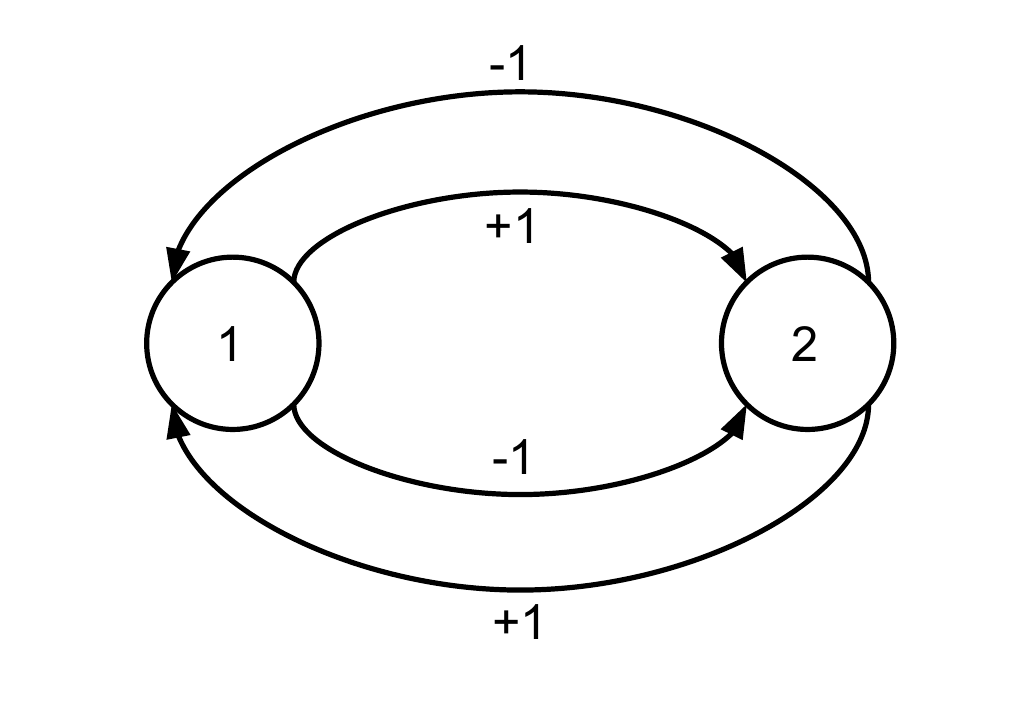}
    \hspace*{-7mm}
    \includegraphics[width=0.55\textwidth]{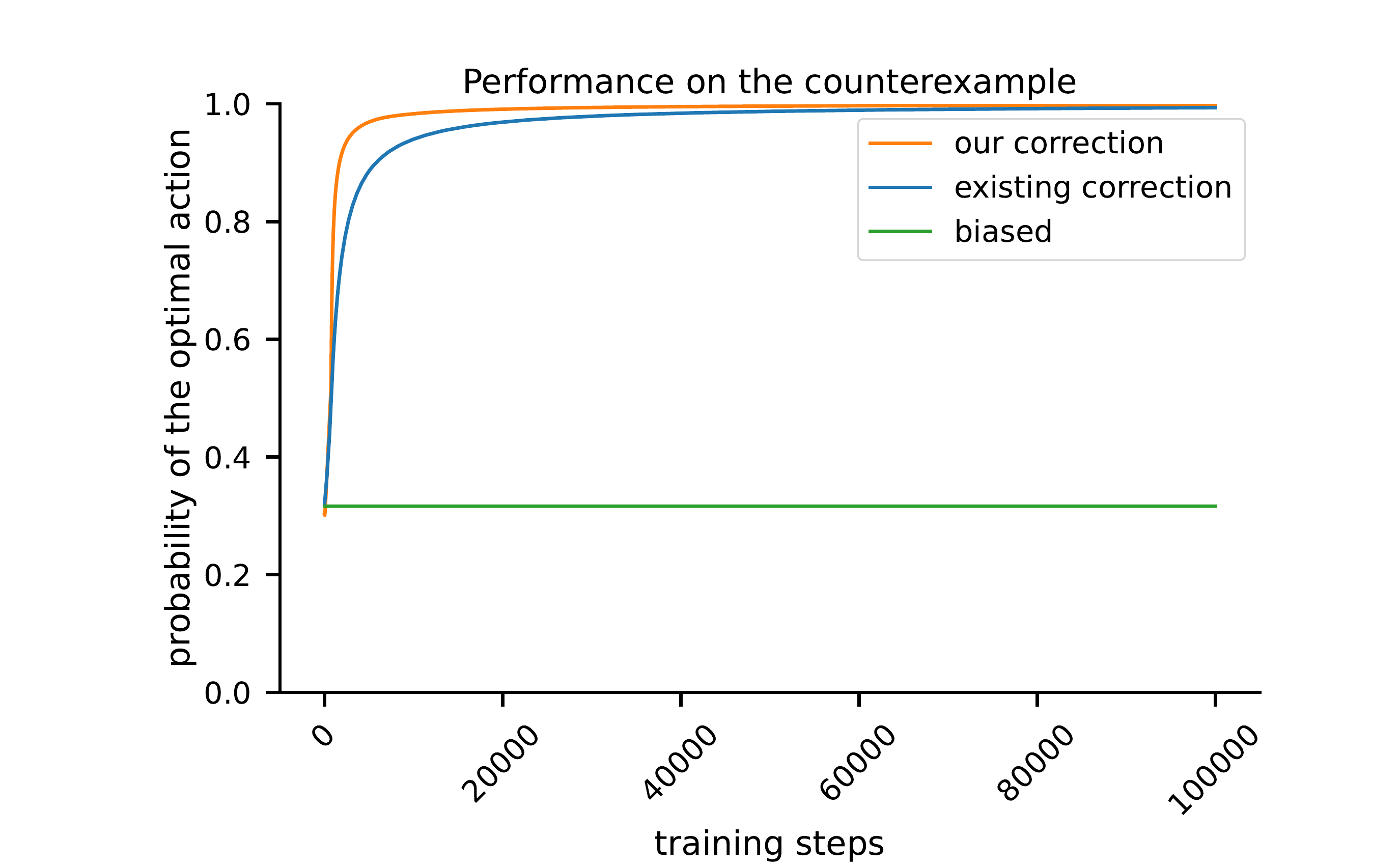}
    \vspace*{-5mm}
    \caption{The top figure shows a deterministic two-state environment where the agent always starts from state $1$ and takes either top or bottom action. The bottom figure presents the learned probability of choosing the optimal action when the agent uses the original biased gradient or gradients with corrections. More details are given in the Experiments section.}
    \label{fig:counterexample}
    \vspace*{-4mm}
\end{figure}


\section{Introduction}
Maximizing discounted cumulative returns is a commonly chosen surrogate objective for reinforcement learning (RL) 
for many well-performing algorithms \cite{sutton2018reinforcement,fujimoto2018addressing,2210.01800}. Popular policy gradient (PG) algorithms maximize this objective by directly searching for an optimal parameterized policy. Although the policy gradient theorem \cite{sutton1999policy} suggests model-free PG algorithms based on the discounted stationary state distribution and the discounted expected returns for estimating gradients, the widely used on-policy algorithms \cite{schulman2017proximal,wu2017scalable} make a biased approximation of the gradients with undiscounted stationary state distribution and discounted returns, leading to the \emph{discount-factor mismatch issue}. This biased gradient may result in a sub-optimal policy \cite{nota2019policy} as shown in Figure \ref{fig:counterexample}, which does not emphasize early rewards and causes degenerate learning behavior. One solution to this issue is including the discount factor's power term $\gamma^t$ in the update \cite{thomas2014bias}, where $t$ is the showing-up time of the state-action pair in a trajectory. Nevertheless, this solution lacks theoretical support for learning from a batch of state-action pairs, and it  can also provide worse performance than the biased gradient \cite{10.5555/3535850.3536016}.

In solving the discount-factor mismatch issue, we propose keeping the \emph{undiscounted-state-sampling setting} with discounted returns of existing on-policy algorithms since data from the undiscounted state distribution is more natural to collect. 
Thus, to correct the bias of gradient estimates, we introduce a distribution correction term, \emph{averaging correction}, under the undiscounted-state-sampling setting. It improves the existing solution by averaging $\gamma^t$ over all steps relevant to a state along the trajectory. Thus, averaging correction efficiently utilizes its information at multiple time steps when sampling a state-action pair. Moreover, our correction is equivalent to the ratio between the discounted and undiscounted state distributions. Therefore, using our correction as a factor under the undiscounted-state-sampling setting gives the true policy gradient, which is stated formally in the \emph{extended policy gradient theorem} in Section 4. After providing mathematical modelling of the existing correction under stationary state distribution, we further find that our correction has a lower variance than the existing one.

We develop a gradient estimator based on the extended policy gradient theorem for on-policy algorithms, by including a novel distribution correction. Our proposed averaging correction is approximated through a regression task, a new and simple way of approximating the state distribution ratio. Our gradient estimator successfully avoids the degenerate policy and converges to the optimal policy on the counterexample in Figure \ref{fig:counterexample}. Moreover, even when our correction is not exactly computed, our estimator still reduces biases from the discount-factor mismatch, shown experimentally on a discrete Reacher task. We further test the performance of our correction applied to other on-policy policy gradient estimators, including batch actor-critic \cite{konda1999actor} and proximal policy optimization (PPO) \cite{schulman2017proximal}. Our work establishes a more principled policy gradient estimator with competitive experimental results. It improves over the performance of uncorrected estimators in tasks with substantial biases arising from the discount-factor mismatch and achieves comparable performance to uncorrected estimators on other discrete and continuous control tasks, distinguishing it from the existing $\gamma^t$ correction.

\section{Background}
\subsection*{Markov Decision Process}
A Markov decision process (MDP) \cite{sutton2018reinforcement} is a tuple $M=\langle \mathcal{S},\mathcal{A},p,\gamma,r,\rho\rangle$ where $\mathcal{S}$ is the state space, $\mathcal{A}$ is the action space, $p$ is the time-homogeneous transition probability distribution denoting $p(s'|s,a)$ as the probability of transitioning to $s'$ from the state-action pair $(s,a)$, $\gamma\in (0,1)$ is the discount factor, $r(s,a)$ is the expected rewards received after state-action pair $(s,a)$, and $\rho$ is the distribution of the initial state. We use finite state and action space in our analysis for simplicity. At each step $t$, the agent applies an action according to a policy $\pi$. Then the agent receives a reward and transits to the next state. In episodic tasks, a trajectory terminates at a termination state, after which the agent transitions to the next state according to the initial distribution. In continuing tasks, a trajectory lasts forever, and the length goes to infinity. Furthermore, our paper adopts unified notation, where episodic tasks have state-dependent discount factors, zero for terminal states and the same constant in $(0, 1)$ for other states. But for simplification, we ignore the dependence on states in notations and write only $\gamma$.

The agent's goal is to find a policy $\pi$ which maximizes returns for a trajectory in expectation. Then a commonly used objective is to maximize the expected discounted cumulative return, denoted by $J(\pi)$ and is defined as 
\begin{equation*}
J(\pi) = (1-\gamma)\E_\pi \left[ \sum_{t=0}^{\infty} \gamma^t r(S_t,A_t) \right],
\end{equation*}
where $S_0 \sim \rho$, $A_t \sim \pi(\cdot|S_t)$ and $S_{t+1} \sim p(\cdot|S_t,A_t)$ for $t \in \mathbb{N}$. The policy $\pi$ in the subscript means that the involved random variables depend on the policy. The action value function for a stationary policy $\pi$, denoted by $q_{\pi,\gamma}$, is defined as the expected return from the state-action pair $(s, a)$ following the policy $\pi$:
\begin{equation*}
q_{\pi,\gamma}(s,a) =
\E_\pi \left[ \sum_{t=0}^{\infty} \gamma^t r(S_t,A_t) \bigg| S_0=s, A_0=a \right].
\end{equation*} 

For policy gradient algorithms, policies are parameterized by $\theta$, written as $\pi_{\theta}$. This paper focuses on a specific parameterization of policies $\pi_{\theta}$ such that the corresponding Markov chain is irreducible and all states are positive recurrent under all possible parameters $\theta$. An irreducible Markov chain satisfies that for any two states, $s$ and $s'$, the probability of transiting from one state to the other is positive at some time step. Moreover, the definition of states' positive recurrence depends on the recurrence time $\tau_s^+(s)$, which represents the time elapsed to revisit a state $s$ and is defined as:
\begin{equation*}
    \tau^+_s(s) = \min \{t>0:S_t =s ,S_0 =s\}.
\end{equation*}
A positive recurrent state has a finite expected recurrence time, that is $\mathbb{E}_\pi[\tau^+_s(s)] < \infty$ for all $s$. Softmax policies with finite parameters usually satisfy the requirements.

\subsection*{Stationary State Distribution}
The undiscounted stationary distribution, denoted by $d_{\pi}$ is defined as the distribution satisfying $\sum_{s\in\mathcal{S}}d_{\pi}(s)p_{\pi}(s'|s)=d_{\pi}(s')$. It has multiple analytical forms stated in Sutton and Barto \citeyearpar{sutton2018reinforcement} and  Grimmett and Stirzaker (\citeyear{grimmett2020probability}, Theorem 6.4.3), summarized in the following lemma; the proof is written in Appendix A.

\begin{lemma}[Forms of Undiscounted Stationary Distribution] \label{lemma:form_stat_dist}
Under the irreducibility of the Markov chain and positive recurrences of all states under all policies $\pi$, we have the following:
  \begin{align*}
      d_{\pi}(s) &= \lim_{T \to \infty} \frac{1}{T} \sum_{t=0}^{T-1} P_{\pi}(S_t=s) = \frac{1}{\mathbb{E}_\pi[\tau^+_s(s)]}.
  \end{align*}
\end{lemma}

The discounted stationary state distribution, denoted by $d_{\pi,\gamma}$, is defined as the distribution satisfying the following equation for all states $s' \in \mathcal{S}$:
\begin{equation*}
\sum_{s\in\mathcal{S}} d_{\pi,\gamma}(s)[\gamma p_{\pi}(s'|s)+(1-\gamma) \rho(s')] = d_{\pi,\gamma}(s') ,
\end{equation*} 
where $p_{\pi}(s'|s) = \sum_{a\in\mathcal{A}}\pi(a|s)p(s'|s,a)$. 
It can be interpreted as the undiscounted stationary distribution of an MDP where the task terminates with probability $1-\gamma$ at each time step. 
A common analytical form of the discounted stationary distribution can be written as
\begin{equation}
d_{\pi,\gamma}(s) = (1-\gamma)\sum_{t = 0}^\infty \gamma^t P_{\pi}(S_t=s).
\label{eq:def_stationary}
\end{equation}

\subsection*{Policy Gradient Theorem}
Policy gradient methods parameterize the policies with a parameter $\theta$ and update it using the gradient of the objective $J(\pi_{\theta})$, called the \emph{policy gradient}. With the help of the policy gradient theorem \cite{sutton1999policy}, the policy gradient $\nabla_{\theta}J(\pi_{\theta})$ can be written as 
\begin{align*}
&\sum_{s\in\mathcal{S}} d_{\pi,\gamma}(s)\sum_{a\in\mathcal{A}}\pi_{\theta}(a|s)\nabla_{\theta} \log\pi_{\theta}(a|s) q_{\pi_{\theta},\gamma}(s,a),
\end{align*}
consisting of the action value, the gradient of the action likelihood, and the discounted stationary distribution. 

This theorem leads to an unbiased gradient estimator, 
\begin{equation}
   \nabla_{\theta} \log\pi_{\theta}(A|S) q_{\pi_{\theta},\gamma}(S,A),
   \label{eq:unbiased_grad_est}
\end{equation}
with state $S\sim d_{\pi_{\theta},\gamma}(\cdot)$ and action $A \sim \pi_{\theta}(\cdot|S)$.

\section{Related Works}
The widely used gradient estimator does not match with the desired estimator in Equation \ref{eq:unbiased_grad_est}. Instead, those estimators are estimating an incorrect gradient with discounted value functions but the undiscounted state distribution, which is
\begin{equation}
\sum_{s\in\mathcal{S}} d_{\pi}(s)\sum_{a\in\mathcal{A}}\pi_{\theta}(a|s)\nabla_{\theta} \log\pi_{\theta}(a|s) q_{\pi_{\theta},\gamma}(s,a).
\label{eq:wrong_grad}
\end{equation}
For example, the proposed gradient of proximal policy optimization (PPO) \cite{schulman2017proximal} requires states sampled from the discounted stationary state distribution. PPO's gradient estimator is shown in the following, which clips the ratio of policy change to avoid rapid change,
\begin{align*}
     \text{\small min} &\{ \nabla_{\theta}e_{(S,A)}(\theta) H_{\pi_{\theta}}(S,A),\\
     &\text{\small clip}(e_{(S,A)}(\theta),1-\epsilon,1+\epsilon) H_{\pi_{\theta}}(S,A)\},
\end{align*}
where $S \sim d_{\pi_{\theta},\gamma}$, $A \sim \pi_{\theta}$, $e_{(s,a)}(\theta) = \cfrac{\pi_{\theta}(a|s)}{\pi_{\theta_{old}}(a|s)}$ is the policy ratio and $H_{\pi_{\theta}}(s,a) = q_{\pi_{\theta}}(s,a) - \mathbb{E}_{A \sim \pi_{\theta}} [ q_{\pi_{\theta}}(s,A)]$ is the advantage value. 
However, the algorithm, with the help of a data buffer $\mathcal{D}$, approximates the undiscounted state distribution. The states sampled from the data buffer are weighted by a sampling distribution $\hat{d}_{\mathcal{D}}$, equaling to
\begin{align}
    \hat{d}_{\mathcal{D}}(s) = \frac{1}{|\mathcal{D}|} \sum_{t=0}^{|\mathcal{D}|-1} \mathds{1}[S_t =s].
    \label{eq:sampling_dist}
\end{align}
As the buffer size reaches infinity, this sampling distribution converges to the undiscounted state distribution. Thus, the gradient estimator is biased under the undiscounted-state-sampling. 
The same issue exists for almost all on-policy policy gradient algorithms, including natural policy gradient (NPG) \cite{kakade2001natural}, trust region policy optimization (TRPO) \cite{schulman2015trust}, ACKTR \cite{wu2017scalable}, and other popular actor-critic-based algorithms \cite{konda1999actor,mnih2016asynchronous, cobbe2021phasic}.

Nota and Thomas \citeyearpar{nota2019policy} have shown that these widely used estimators under the undiscounted-state-sampling setting are not a gradient of any function. Additionally, they show that this mismatch issue may cause a suboptimal policy no matter whether the objective is the average reward or the discounted one. An existing correction includes the discount factor's powers $\gamma^t$ in the update as a factor. \citet{thomas2014bias} and REINFORCE \cite{williams1992simple} deal with a full trajectory of data and multiply the power term at each time step. But the trajectory-based update is not data efficient. \citet{10.5555/3535850.3536016} and incremental actor-critic \cite{sutton2018reinforcement} deal with state-action pairs with each state drawn from an approximated undiscounted stationary distribution and multiply the power term to each data transition. However, they do not show analytically that this modification leads to an unbiased estimate, which is completed in our paper in Section 4. Moreover, \citet{10.5555/3535850.3536016} test the correction experimentally on PPO, where its performance decreases compared to the original biased gradient in a few tasks. It is not unexpected since states that appear later in episodes can be similar to early states and will be underutilized since the existing correction assigns tiny weights to them. Thus, we integrate a state's information over the trajectory, leading to our averaging correction. 

The discount-factor mismatch issue can also be seen as a state distribution correction problem, which has been studied under off-policy settings. Emphatic weighting \cite{imani2018off,jiang2022learning,mahmood2015emphatic} develops a state weighting to match the excursion objective, where the agent changes to the target policy from the undiscounted stationary distribution of the behaviour policy. But the emphatic weighting cannot work for the discounted objective. On the other hand, \citet{liu2019off} and \citet{liu2018breaking} find that the state distribution ratio $\frac{d_{\pi,\gamma}}{d_{\mu,\gamma}}$ can be approximated by solving the following problem for $w$:
\begin{align*}
    &\gamma \mathbb{E}_{(S,A,S') \sim (d_{\pi},\pi,p)}[(w(S) \frac{\pi(A|S)}{\mu(A|S)}-w(S'))f(S')]\\
    &+(1-\gamma) \mathbb{E}_{S \sim \rho}[(1-w(S))f(S)] = 0,
\end{align*}
for all measurable functions $f$. A form to correct our issue is not developed in the original paper but is doable. However, it results in a harder optimization problem than our regression task and requires knowledge of the initial state distribution. Moreover, COP-TD \cite{gelada2019off,hallak2017consistent} induces a Bellman-equation-like form of state distribution ratio and updates the ratio incrementally. But this technique requires information on both the undiscounted stationary state distribution and the initial state distribution and thus cannot be generally used.

\section{Extended Policy Gradient Theorem}
To correct discount-factor mismatch and while sampling from the undiscounted state distribution, we need a state distribution correction term. An existing correction multiplies the discount factor's powers into the gradient but is only proven to be correct in utilizing trajectory data. In this paper, we statistically model the discount factor's power under stationary state distribution. Next, we introduce our correction term, which averages the discount factor's powers of a state along the trajectory, and present the extended policy gradient theorem. At last, we discuss the relationship between the existing correction and our averaging correction.

\subsection*{An Existing Correction}
Let us start by rigorously modelling the existing correction $\gamma^t$ under stationary state distribution. When a state is sampled from the undiscounted state distribution, 
the power term $t$ is not a sequential step but a random variable. 
What distribution does this power term follow? We model this power term as a random variable $K_T(s)$, dependent on a length variable $T$. Given a trajectory $\tau$ of length $T$, a state $s$ may show up multiple times and the power term is sampled uniformly among times at which state $s$ is visited. Therefore, the distribution of this random variable is defined as
\[P_\pi(K_T(s) = t) = \begin{cases}
\E_{\tau\sim\rho,\pi}&\Big[ \frac{\mathds{1}[S_t=s]}{\sum_{k=0}^{T-1}\mathds{1}[S_k=s]} \Big], \\
&\forall t=0,1,\cdots,T-1,\\
0, &\forall t \ge T.
\end{cases}\]
Here, the expectation is taken over all possible trajectories $\tau$. Then, a gradient estimator with the existing correction under stationary distribution is 
\begin{align*}
    \E_{S\sim d_{\pi}}\left[  \gamma^{K_T(S)} \sum_{a\in\mathcal{A}}\nabla \pi(a|S) q_{\pi,\gamma}(S,a)\right],
\end{align*}
which is the estimator used by \citet{10.5555/3535850.3536016}. There are some obvious issues with the above naive correction. First, it adds variance in gradient estimates since the algorithm has to sample further a power term. Second, it decays exponentially as the time step increases, wasting samples showing up late in an episode.

\subsection*{Our Proposed Averaging Correction}
To alleviate the issues of the existing correction, we propose to average the related $\gamma^t$ terms of a state along the trajectory. Let us consider a trajectory $\tau$ until the time step $T$, where a state $s$ has shown up multiple times. Instead of uniformly sampling a time step $t$ at which $s$ is visited, we count in all the information of a state $s$ along the trajectory and average all related $\gamma^t$ when $s$ shows up. This averaging term for each state $s$ can be expressed mathematically as 
\begin{align*}
c_T(s, \tau) &=  \frac{ \sum_{t = 0}^{T-1} \gamma^{t} \mathds{1}[S_{t} = s] }{ \sum_{k = 0}^{T-1} \mathds{1}[S_{k} = s] }.
\end{align*}
The numerator is the summation of all discount factor's powers for time steps where state $s$ shows up, and the denominator is the number of showings, which together make up the averaging.

Furthermore, we extend the averaging term to a distribution corrector and develop an extended policy gradient theorem under the undiscounted-state-sampling setting. The proposed distribution correction is independent of any specific trajectory $\tau$ after taking expectations and becomes independent of the episode length variable $T$ as it approaches infinity. The distribution correction $c_{\pi}(s)$, named \emph{averging correction}, is defined as
\begin{align}
    c_\pi(s)&:= \lim_{T \to \infty} c_{\pi,T}(s),    \label{def::avg correction}
\\
    c_{\pi,T}(s) &= \E_{\tau\sim\rho,\pi} \left[(1-\gamma) T c_T(s, \tau)\right]\label{def::finite avg correction}.
\end{align}
First, we present that the averaging correction exists after taking the limit and is finite for all positive recurrent states, stated in Lemma \ref{lemma:C_existance}. The proof follows the ergodic theorem \cite{norris1998markov} and is given in Appendix B.

\begin{lemma}\label{lemma:C_existance}
For any positive recurrent state $s$, the averaging correction $c_{\pi}(s)$ exists and is bounded by the expected recurrence time:
\begin{equation*}
    c{_\pi}(s) = \mathbb{E}_\pi[\tau^+_s(s)] d_{\pi,\gamma}(s)\le \mathbb{E}_\pi[\tau^+_s(s)] < \infty.
\end{equation*}
\end{lemma}

Next, we prove that multiplying our averaging correction under the undiscounted-state-sampling setting gives the correct policy gradient and thus gives rise to unbiased gradient estimators, shown in the following theorem. The proof is straightforward; Lemma \ref{lemma:form_stat_dist} tells that the expected recurrence time equals the reverse of the undiscounted distribution. Therefore, combining with results  in Lemma \ref{lemma:C_existance}, our correction equals the distribution ratio, $\frac{d_{\pi,\gamma}(s)}{d_{\pi}(s)}$ and thus, the bias from the discount-factor mismatch issue is overcome. Formal proof is given in Appendix B.

\begin{theorem}[Extended Policy Gradient Theorem]\label{thm:averaging-correction-PGT}
Given an irreducible Markov decision process with positive recurrent states, the policy gradient can be expressed as
\begin{align}
\nabla J(\pi)&= \mathbb{E}_{S\sim d_{\pi,\gamma}}[\sum_{a \in\mathcal{A}}\nabla \pi(a|S)q_{\pi,\gamma}(S,a)] \\
&= \mathbb{E}_{S \sim d_{\pi}}[c_\pi(S)\sum_{a \in\mathcal{A}}\nabla \pi(a|S)q_{\pi,\gamma}(S,a)]. \label{eq:unbiased_modified_gradient}
\end{align}
\end{theorem}

\subsection*{Relation between Two Corrections}
It turns out that the finite term $c_{\pi,T}(s)$ is the expected value of existing $\gamma^{K_T(s)}$ up to a constant $T(1-\gamma)$.
\begin{align*}
    &T(1-\gamma) \E_\pi[\gamma^{K_T(s)}] \\
    & = T(1-\gamma) \sum_{t=0}^{T-1} \gamma^t P_\pi(K_T(s) =t)\\
    &= T(1-\gamma) \E_\pi\left[\frac{\sum_{t=0}^{T-1} \gamma^t\mathds{1}[S_t=s]}{\sum_{k=0}^{T-1}\mathds{1}[S_k=s]}\right] \\
    &=c_{\pi, T}(s).
\end{align*}
The second equality is gained by substituting in the probability mass function of $K_T(s)$. Then, taking limits on both sides gives us
\begin{align*}
    c_\pi(s) = \lim_{T \to \infty} T(1-\gamma)\E_\pi[\gamma^{K_T(s)}].
\end{align*} 
Notice that our correction integrates information of a state over the trajectory and turns out to be the mean of the existing correction, and hence with less variance. 

Moreover, the correctness of the existing correction follows from the extended policy gradient theorem, which is shown in the following corollary and proved in Appendix B. Thus, our paper for the first time proves that the existing solution by discount factor's powers fixes the bias.

\begin{corollary}\label{thm:naive_modified_PGT}
Given an irreducible Markov decision process with positive recurrent states, the gradient of the discounted objective can be written as:
\vspace*{0mm}
\begin{align*}
    &\scalebox{0.98}{$\nabla J(\pi)= \mathbb{E}_{S\sim d_{\pi,\gamma}}[\sum_{a \in\mathcal{A}}\nabla \pi(a|S)q_{\pi,\gamma}(S,a)]$}\\
    \vspace*{0mm}\\
    &=  \scalebox{0.9}{$\Lim{T \to \infty} \E_{S \sim d_{\pi}}\left[ T(1-\gamma)\gamma^{K_T(S)} \sum_{a\in\mathcal{A}}\nabla \pi(a|S) q_{\pi,\gamma}(S,a)\right]$}.
\end{align*}
\end{corollary}

\section{Our Proposed Algorithm}
The extended policy gradient theorem in Equation \ref{eq:unbiased_modified_gradient} induces an unbiased gradient estimator
\begin{equation}
    c_{\pi_{\theta}}(S) \nabla_{\theta} \log \pi_{\theta}(A|S) q_{\pi_{\theta},\gamma}(S,A),
    \label{eq:modified_gradient_estimator}
\end{equation}
where $S \sim d_{\pi_{\theta}}$ and $A \sim \pi_{\theta}$. We propose to modify current on-policy policy gradient estimators according to this unbiased estimator, by adding our averaging correction under the undiscounted-state-sampling setting. We start by estimating our averaging correction. Then we present how to include our gradient estimators into current algorithms, such as the batch actor-critic (BAC) and PPO.

\subsection*{Practical Approximated Correction}
Our correction, $c_{\pi}(s)$ serves as the state distribution ratio, which is hard to compute analytically. However, our form of writing it as an average gives a new way for estimation through regression. First, our averaging correction can be approached by $c_{\pi,T}(s)$ in Equation \ref{def::finite avg correction} using finite time steps. Then, the expectation can be evaluated on a data buffer $\mathcal{D}$, consisting of $k$ trajectories of length $T$ under the policy $\pi$, as the following:
\begin{align*}
    \hat{c}_{\mathcal{D}}(s) = (1-\gamma) T \frac{\sum_{i=1}^{|\mathcal{D}|} \gamma^{t_i} \mathds{1}[S_{i} =s]}{\sum_{i=1}^{|\mathcal{D}|} \mathds{1}[S_{i} =s]}.
\end{align*}

This buffer-based approximation can help correct the state distribution. Originally, states are sampled from the sampling distribution $\hat{d}_{\mathcal{D}}(s)$. Then states are reweighted by multiplying our approximated distribution correction, leading to a corrected state distribution with probabilty density function $c_{\mathcal{D}}(s)\hat{d}_{\mathcal{D}}(s)$. As shown in the following proposition, this corrected state distribution converges to the discounted state distribution at the rate of $O(\max\{\sqrt{\frac{T}{|\mathcal{D}|}},\gamma^T\})$. More detailed explanations are written in Appendix C.

\begin{proposition}
Given a data buffer $\mathcal{D}$ collected on an irreducible and positive recurrent MDP under policy $\pi$, consisting of $k$ trajectories, each with length $T$, if $$k \ge \frac{2}{\epsilon^2}\log\frac{|\mathcal{S}|}{\delta} \text{ and } T \ge \cfrac{\log \frac{\epsilon}{2}}{\log \gamma},$$ then with probability at least $1-\delta$, we have
\begin{align*}
    \max _{s\in\mathcal{S}}|\hat{d}_{\mathcal{D}}(s) c_{\mathcal{D}}(s) - d_{\pi,\gamma}(s)| \le \epsilon.
\end{align*}
\end{proposition}

Thus, the accuracy of the approximation $\hat{c}_{\mathcal{D}}(s)$ is expected to improve when the trajectory length $T$ and the number of trajectories $k$ increase. Hence, a larger data buffer may enhance the distribution correction but lower the data efficiency. Therefore, the size of the buffer is treated as a hyperparameter during training.

\begin{algorithm}[tb]
   \caption{BAC with Averaging Correction}
   \label{alg:averaging_pg}
\begin{algorithmic}
   \STATE {\bfseries Initialize:} Parameters $\theta_0$ for policy network, $w_0$ for value network, and $\sigma_0$ for correction network, Learning rates $\alpha_1$, $\alpha_2$, and $\alpha_3$
   \STATE{Start the task and get initial state} $S_0 \sim \rho$
   \FOR{learning step $k=0,1,...$}
   \FOR{time step $t= 0,1,...$}
   \STATE{Sample action} $A_t \sim \pi_{\theta_k}(\cdot|S_t)$ and play\\
   Get reward $R_{t+1}$ and next state $S_{t+1}$\\
   \IF{$S_{t+1}$ is a terminal state}
   \STATE{Restart the episode and get initial state} $S_0$
   \ENDIF\\
   Store $(S_t,A_t,R_{t+1},t)$ into the buffer
   
   \IF{There are $|\mathcal{D}|$ transitions in the buffer}
   \STATE{\bfseries Update correction network parameter}
   \STATE $\sigma_{k,0} \leftarrow \sigma_{k}$ \\
   \FOR{step $m =1,...,M$}
   \STATE \small{$\sigma_{k,m} \leftarrow \sigma_{k,m-1} - \alpha_1 \nabla \mathcal{L}_{correction}(\sigma_{k,m-1})$}\\
   \ENDFOR\\
   \STATE $\sigma_{k+1} \leftarrow \sigma_{k,M}$ \\
   \STATE{\bfseries Update policy network parameter}\\
   \STATE $\theta_{k+1}\leftarrow \theta_k + \alpha_2 \widehat{\nabla J(\pi_{\theta_k})};$\quad using (\ref{eq:grad_est}) \\
   \STATE{\bfseries Update value network parameter}\\
   \STATE $w_{k+1}\leftarrow w_k - \alpha_3 \nabla \mathcal{L}_{value}(w_k)$\\
   \STATE{Empty the buffer}
   \ENDIF\\
   \ENDFOR \\
   \ENDFOR
\end{algorithmic}
\end{algorithm}

This averaging term, $\hat{c}_{\mathcal{D}}(s)$, is further estimated by a neural network with parameter $\sigma$ and output $f_{\sigma}$. This avoids counting for each state, which is computationally prohibitive for a large state space. The neural network takes a state $s$ as an input and its discount factor's power $\gamma^t$ as its target, with $t$ being its step stored in the buffer. The learning objective under each buffer is to minimize mean square error between the output $f_{\sigma}(s)$ and the discount factor's power, that is 
\begin{align*}
    \mathcal{L}_{correction}(\sigma) = \sum_{i=1}^{|\mathcal{D}|} \frac{1}{|\mathcal{D}|}(f_{\sigma}(S_i)-\gamma^i)^2.
\end{align*}
The optimal solution for the above objective is the averaging of the discount factor's powers, $\hat{c}_{\mathcal{D}}(s)$, up to a constant term. 

Approximating our averaging correction adds bias to the gradient estimator, but our algorithm still lessens the bias from the mismatching state distribution, shown experimentally in Section 6. Moreover, our estimated averaging correction also has less variance than the existing correction since ours averages multiple samples of a state instead of depending on only one sample.

\subsection*{Batch Actor-Critic with Averaging Correction}
In batch actor-critic, a buffer $\mathcal{D}$ collects multiple transitions, forms them into one batch and makes a single update, guaranteeing an on-policy update. As discussed, we correct the previous state weights of BAC, the sampling distribution $\hat{d}_{\mathcal{D}}$, by multiplying our averaging correction.

Finally, our proposed policy gradient estimator, denoted by $\widehat{\nabla_{\theta} J(\pi_{\theta})}$, includes an estimated correction to reemphasize states and is proportional to
\vspace*{-2mm}
\begin{align}
    \frac{1}{D}\sum_{i=1}^{|\mathcal{D}|} f_{\sigma}(S_i) \nabla_{\theta} \log \pi_{\theta}(A_i|S_i) H_{w}(S_i,A_i),
    \label{eq:grad_est}
\end{align}
where $H_{w}(S_i,A_i)$ is the estimated advantage value, equaling to $r(S_i,A_i) + \gamma V_w(S_{i+1}) - V_w(S_{i})$. Another neural network estimates value functions with parameter $w$, denoted by $V_{w}$, under the mean squared temporal difference loss, denoted by $\mathcal{L}_{value}(w)$. 

The algorithm is shown in Algorithm \ref{alg:averaging_pg}. The agent executes a policy ${\pi_{\theta_k}}$ and collects a buffer of data. Then, after learning the averaging correction, the agent updates the value function parameter and the policy parameter by our modified gradient estimator in Equation \ref{eq:grad_est}. 

\subsection*{Add-on Correction Algorithm}
Any on-policy policy gradient algorithm can multiply our approximated averaging correction at the cost of one more neural network to learn it and gain a less biased gradient estimator. Our paper introduces the modified PPO gradient estimator as an example.

The modified PPO adds a correction network and includes our correction in the modified clipped objective: 
\begin{align*}
     \frac{1}{|\mathcal{D}|} \sum_{i=1}^{|\mathcal{D}|} f_{\sigma}(S_i) \text{\small min} \{e_i(\theta) H_i, \text{\small clip}(e_i(\theta),1-\epsilon,1+\epsilon) H_i\},
\end{align*}
where $e_i(\theta) = \cfrac{\pi_{\theta}(A_i|S_i)}{\pi_{\theta_{old}}(A_i|S_i)}$ and $H_i$ is the learnt generalized advantage estimation for state-action $(S_i,A_i)$. 



\section{Experiments}
\begin{figure}
    \centering
    \hspace*{-11mm}
    \includegraphics[width=0.57\textwidth]{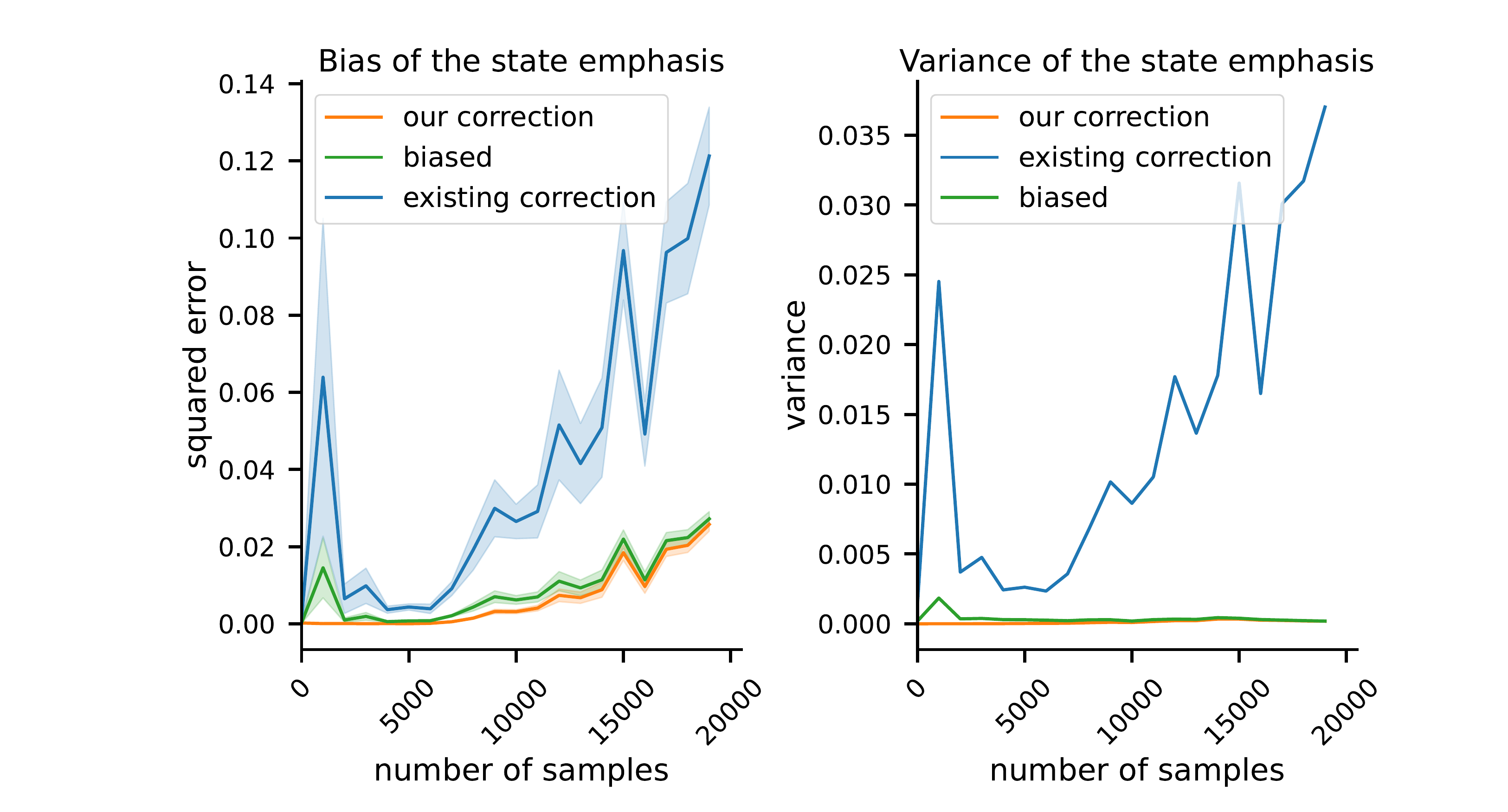}
    \vspace*{-3mm}
    \caption{On the left, we show the bias of the state emphasis used by the three gradient estimators. It is the sampling distribution estimated by the buffer for the biased gradient and the correction times the sampling distribution for the other two corrected estimators. On the right, we show their variances.}
    \label{fig:bias_ratio}
    \vspace*{-3mm}
\end{figure}
\subsection*{Counterexample}
We test if our estimated correction can avoid the degenerate policy caused by the incorrect gradient under the discount-factor mismatch. The counterexample in Figure \ref{fig:counterexample} is a deterministic and continuing two-state environment. The agent always starts from state $1$ and takes one of the two actions, top and bottom. No matter which action is taken, the agent transits to the other state and receives a reward as labelled in the figure. Notice that the rewards in the two states are the reverse. In the function approximation case, these two states can be indistinguishable and share one policy parameter $\theta\in\mathbb{R}$. Then, the policy always chooses the top action with probability $\pi_{\theta}(top)=\frac{e^{\theta}}{1+e^{\theta}}$. 
\begin{figure*}[tb]
    \centering
    \vspace*{-2mm}
    \includegraphics[width=0.95\textwidth]{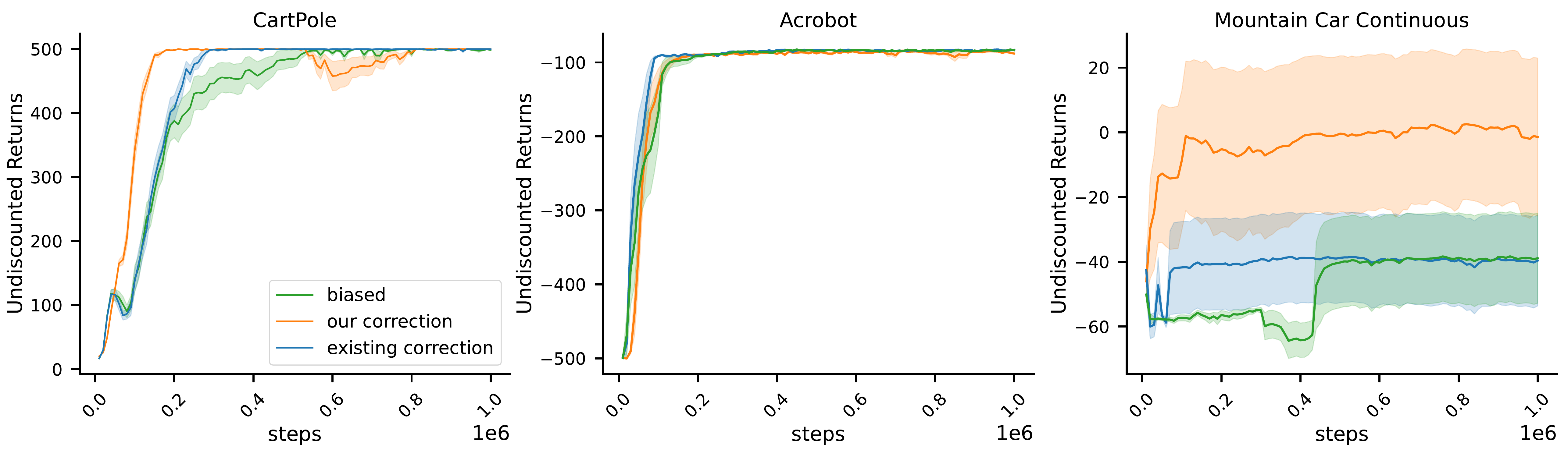}
    \vspace*{-2mm}
    \caption{Performance of modified BAC and two baselines on classic control tasks averaged over $10$ runs.}
    \label{fig:classic_control}
\end{figure*}

\begin{figure*}[htb]
    \centering
    \hspace*{-2mm}
    \includegraphics[width=1.0\textwidth]{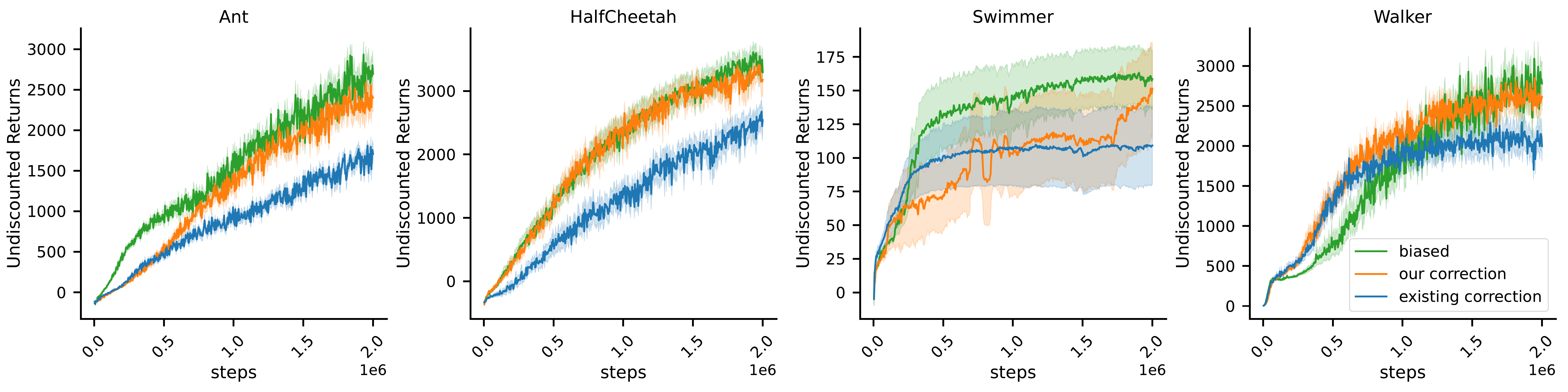}
    \vspace*{-2mm}
    \caption{Performance of modified PPO and two baselines on MuJoCo tasks averaged over $10$ runs.}
    \label{fig:mujoco}
    \vspace*{-3mm}
\end{figure*}

When we want to maximize a discounted objective with a discount factor smaller than one, the agent cannot learn using the original incorrect gradient shown in Equation \ref{eq:wrong_grad}, since it always equals zero. Specifically, Q-values for the two states are the reverse of each other due to the reverse relationship of expected rewards, and thus
\begin{align*}
    q_{\pi_{\theta},\gamma}(1,top) &= - q_{\pi_{\theta},\gamma}(2,top),\\
    q_{\pi_{\theta},\gamma}(1,bottom) &= -q_{\pi_{\theta},\gamma}(2,bottom).
\end{align*}
The undiscounted stationary state distribution is $d_{\pi_{\theta}}(1) = d_{\pi_{\theta}}(2) = \frac{1}{2}$ since two states show up equally many times in an infinite episode.
In this case, the terms in the incorrect gradient in Equation \ref{eq:wrong_grad} cancel each other, and the agent always receives a zero gradient for all policies. 

Next, we train three batch actor-critic agents with or without corrections, all using true Q-values. The learnt policies are shown at the bottom of Figure \ref{fig:counterexample} with the discount factor equaling $0.9$. More results under several discount factors are shown in Appendix D, and the learning curves are similar. We plot the learnt probability of choosing the optimal action along the training steps. While the original gradient returns a deficient policy, the corrected gradients converge to the optimal policy. Our correction coloured in orange, also improves the learning speed. Even if our approximated averaging correction does not fully correct the bias due to its approximation error, it can appropriately weigh states and fix the suboptimal policy in certain cases.

\subsection*{Bias Reduction Analysis}
We design experiments to study whether our algorithm with an approximated averaging correction can still reduce the bias caused by the discount-factor mismatch. We consider the bias as the difference between the true discounted state distribution and the state weighting of each algorithm. The original algorithm weighs states according to the sampling distribution $\hat{d}_{\mathcal{D}}$ from a data buffer $\mathcal{D}$ in Equation \ref{eq:sampling_dist}. The existing and our correction reweigh states and multiply the sampling distribution by the corresponding correction. Moreover, we analyze the variance of the state weightings over multiple data buffers. Here, we study three agents: the original biased batch actor-critic, the batch actor-critic with the existing correction, and the batch actor-critic with estimated averaging correction.

We conduct our bias analysis on a discrete Reacher environment where the knowledge of the dynamics transition matrix is known. Thus the discounted state distribution can be analytically calculated. The environment consists of a $9 \times 9$ grid with the 2D coordinates of cells as states. The agent has eight actions, moving to any adjacent cells. In each episode, the agent is initialized randomly and always gains a penalty except at the center. The episode terminates after $500$ steps. If the agent reaches the center before termination, it is set to a state randomly regardless of its action.

On the left, Fig.\ \ref{fig:bias_ratio} shows the squared biases of state emphasis for batch actor-critic algorithms with and without corrections; on the right, it shows the variances. The plots show that our correction in orange has less bias and variance than the existing correction. This low variance of our algorithm is expected since our correction is a state-dependent term and averages the power term $\gamma^t$ over multiple trajectories when the buffer size is large enough. Meanwhile, it also reduces the biases from the original gradient estimator, taking the approximation of the averaging correction into account. Though it is not apparent, our correction reduces a large amount of the bias from the original algorithm as shown more clearly in Fig.\ \ref{fig:bias_ratio_shared} in Appendix E. 

\subsection*{Batch Actor-Critic with Averaging Correction}
Here, we analyze the performance of the modified batch actor-critic (BAC) with our correction and compare it with the original biased BAC and BAC with the existing correction. The performance is tested on three classic-control tasks in the OpenAI gym simulator \cite{1606.01540}. The states of these tasks are the agent's positions and velocities, and the
actions are the torques applied to the agent. The agent is rewarded according to its position. The hyperparameters are shown in Appendix F.

Here, we present undiscounted returns, which are the true objective for OpenAI tasks. But as shown by Zhang et al. (2022) that learning under the total reward gives poor performance. Thus, discounted returns are used to subrogate thanks to well-developed algorithms for the discounted objective. Therefore, we train agents under the discounted objective, treating discount factors as hyperparameters but report undiscounted returns. The performance under the discounted returns is also reported in Appendix H. The learning trends are similar to the results shown above under the undiscounted returns.


Learning curves in Fig. \ref{fig:classic_control} show the total reward averaged across ten random seeds with standard deviation errors on CartPole, Acrobot and continuous MountainCar. All three algorithms show similar performances on Acrobot. But BAC with our correction in orange dominates CartPole and continuous MountainCar with faster learning. The improved final performance on MountainCar further proves that the gradient estimator with our correction can gain better policies due to bias reduction.

\begin{figure*}[htb]
    \centering
    \vspace*{2mm}
    \begin{minipage}{0.38\textwidth}
     \hspace*{-14mm}
     \includegraphics[width=1.45\linewidth]{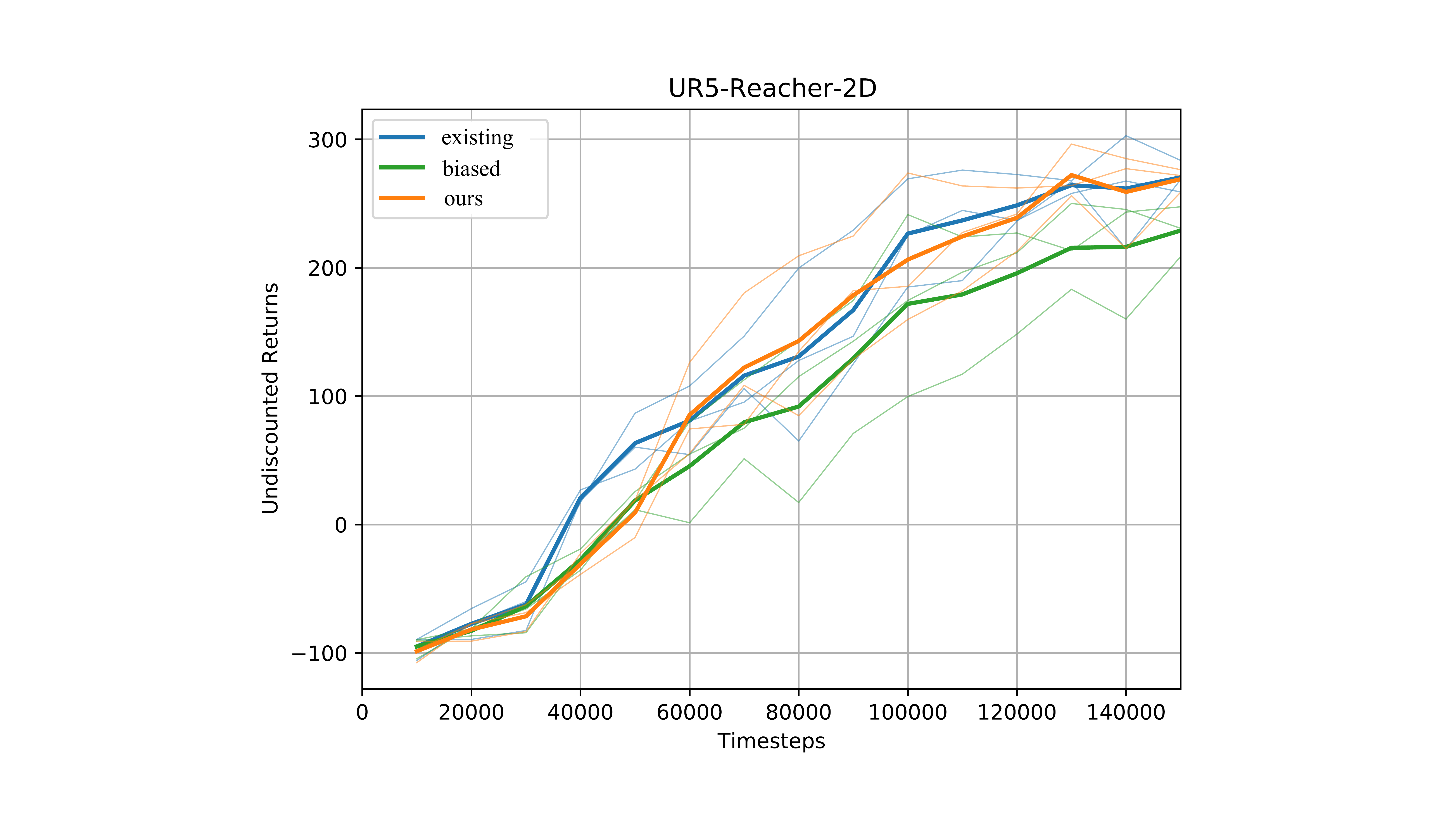}
     \vspace*{-4mm}
     \caption{Performance of modified PPO on UR5.}\label{Fig:ur5}
   \end{minipage}\hfill
    \begin{minipage}{0.62\textwidth}
     \centering
     \includegraphics[width=\linewidth]{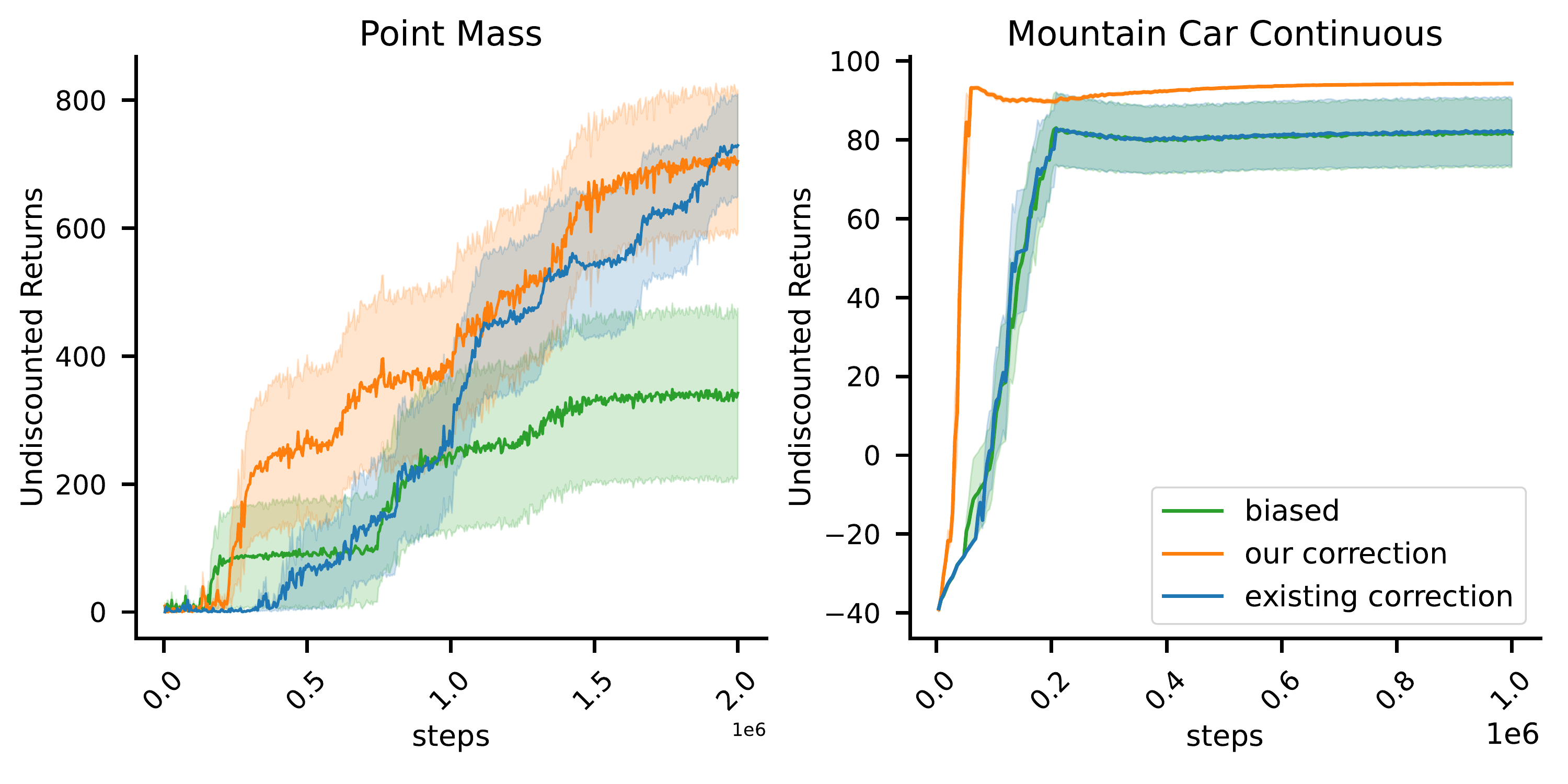}
     \vspace*{-4mm}
     \caption{Performance of modified PPO on Mountain Car Continuous and Point Mass.}\label{Fig:ppo_good_performance}
   \end{minipage}
    \hspace*{-2mm}
\end{figure*}

\subsection*{PPO Add-on Results}
We compare modified PPO with our averaging correction to the original PPO and PPO with the existing correction on four simulated robotic tasks using the MuJoCo simulator \cite{todorov2012mujoco}, two specifically chosen control tasks \cite{tassa2018deepmind,1606.01540} and a real robotic arm task. The original PPO algorithm follows the implementation by OpenAI spinningup \cite{SpinningUp2018}, and the other two algorithms are adjusted based on this implementation. While these algorithms adopt some default values of hyperparameters, several important hyperparameters are tuned for each algorithm on each task. More details are given in Appendix G. Meanwhile, we also report the performance evaluated under the discounted returns in Appendix H for several tasks. The learning trends are similar to the above results under the undiscounted returns.

Figure \ref{fig:mujoco} presents the learning curves on MuJoCo tasks for up to 2M steps averaged over ten random seeds, shown with standard errors by the shaded region. The existing correction hinders the final performance of the agent, as shown in blue. It causes worse returns than the original algorithm. However, PPO with our averaging correction in orange successfully matches the biased algorithm's performance and illustrates that our averaging correction is a better state distribution correction in the sense of performance. 

Meanwhile, we test our algorithm on a real robot task. We train a UR5 robotic arm on the UR-Reacher-2 task, developed by Mahmood et al. (2018). More details of the task and training are shown in Appendix G.3. The performance of our modified PPO and two baselines on the UR-Reacher-2 task is shown in Fig.\ \ref{fig:ur5}. Both modified PPOs with corrections behave slightly better than the original PPO, with faster learning and higher returns.

Fig.\ \ref{Fig:ppo_good_performance} shows the result on Mountain Car Continuous and Point Mass. Here, our modified PPO shows an obvious advantage over the original PPO. This better performance can be because these two tasks have more significant biases in stationary state distributions when ignoring the discount factor. Thus, a correction helps reduce significant bias and improves performance. To support our hypothesis, we should compare the distance between the discounted and the undiscounted state distributions on all tasks. But these distributions cannot be analytically computed. Instead, we measure the max-norm distance between the initial and the sampling state distribution. As proven in Appendix G.4, a large distance between the initial and the undiscounted distributions is a necessary condition for a task to suffer from large bias in the discount-factor mismatch. Our paper uses the sampling distribution to approximate the undiscounted stationary distribution and thus computes the distance between the initial and the sampling distributions as a guesstimate for the bias.

We present the distances in Table \ref{tab:diff_dist}, which are calculated for all policies learnt during training and then averaged. It turns out that these two tasks have more considerable distances than others, implying larger biases from the discount-factor mismatch. It supports our hypothesis that our modified PPO performs better than the original PPO when a missing discount factor in the state distribution gives a larger bias. Notice that the task Swimmer also has a comparably significant distance, but our algorithm does not outperform the original PPO in 2M steps. Perhaps the training step is not long enough. Or a large distance is not sufficient to imply the dominant performance of our algorithm.

\begin{table}[t]
    \centering
    \small
    \vspace*{2mm}
    \begin{tabular}{|c|c|}
        \hline
         Task Name & Averaged Distribution Distance  \\
        \hline
         Walker & 0.0641 \\
         \hline
         Ant & 0.0608\\
         \hline
         HalfCheetah & 0.0952\\
         \hline
         Swimmer & 0.2817\\
         \hline
         Point Mass & 0.2712\\
         \hline
         Mountain Car Continuous & 0.5552\\
         \hline
    \end{tabular}
    \vspace*{2mm}
    \caption{This table shows the averaged maximum norm distance between the initial and sampling state distributions along the training. A large distance may imply a significant bias from the undiscounted state distribution.}
    \label{tab:diff_dist}
    \vspace*{-4mm}
\end{table}



\section{Conclusion}
We proposed averaging correction to address the discount-factor mismatch and utilized it to derive a correct policy gradient, which is summarized as the extended policy gradient theorem.
Our averaging correction provides a new way of estimating the density ratio between two state distributions through a regression task and enables the development of unbiased policy gradient estimators. 
Furthermore, we developed an add-on algorithm to include our averaging correction in any on-policy PG algorithms. We showed that our derived correction was better than the existing correction, with lower variance and more promising learning performance. 
Moreover, our add-on algorithms performed better than the widely-used uncorrected algorithms in tasks where the biases arising from the discount-factor mismatch are large. Meanwhile, they give comparable empirical results to the uncorrected algorithms across all other tasks, establishing well-performing on-policy policy gradient methods that are also technically correct.

\section*{Acknowledgements}
We gratefully acknowledge funding from the CCAI Chairs program, the RLAI laboratory, Amii, and NSERC of Canada. We also thankfully acknowledge the donation of the UR5 arm from the Ocado Group. We would like to thank Jiamin He, Shivam Garg, Samuele Tosatto, Huizhen Yu, Richard Sutton, and Martha White for their valuable discussions.

\bibliography{bib}
\bibliographystyle{icml2022}

\newpage
\appendix
\onecolumn

\section{Proof of Forms of Undiscounted Stationary Distribution}

\begin{lemma}[Forms of Undiscounted Stationary Distribution] 
Under the irreducibility of the Markov chain and positive recurrences of all states under all policies, we have for any policy $\pi$,
  \begin{align*}
      d_{\pi}(s) &= \lim_{T \to \infty} \frac{1}{T} \sum_{t=0}^{T-1} P_{\pi}(S_t=s),\\
      &= \frac{1}{\mathbb{E}[\tau^+_s(s)]}.
  \end{align*}
\end{lemma}

\begin{proof}
First, let us check whether the limit distribution is a stationary distribution. Let us check if it satisfies the definition of the stationary distribution.

\begin{align*}
    \sum_{s \in \mathcal{S}}\lim_{T \to \infty} \frac{1}{T} \sum_{t=0}^{T-1} P_{\pi}(S_t=s) P_{\pi}(s,s')&= \lim_{T \to \infty} \frac{1}{T} \sum_{t=0}^{T-1} P_{\pi}(S_{t+1}=s'),\\
    &= \lim_{T \to \infty} \frac{1}{T} \sum_{t=1}^{T-1} P_{\pi}(S_{t}=s') + \lim_{T \to \infty} \frac{P_{\pi}(S_0=s')- P_{\pi}(S_T=s')}{T}, \\
    & = \lim_{T \to \infty} \frac{1}{T} \sum_{t=0}^{T-1} P_{\pi}(S_{t}=s').
\end{align*}
Thus, $d_{\pi}(s) = \lim_{T \to \infty} \frac{1}{T} \sum_{t=0}^{T-1} P_{\pi}(S_t=s)$.

Next, we show that the stationary distribution is unique and always has the form $\frac{1}{\mathbb{E}[\tau^+_s(s)]}$. For any state $s$, we have
\begin{align*}
    d_{\pi}(s) \E[\tau^+_s(s)] &= d_{\pi}(s) \sum_{t \ge 0} P_{\pi}(\tau^+_s(s)>t |S_0=s),\\
     &= \sum_{s'} d_{\pi}(s')\sum_{t \ge 0} P_{\pi}(\tau^+(s)>t| S_0=s') -\sum_{s' \neq s} d_{\pi}(s') \sum_{t \ge 0} P_{\pi}(\tau^+(s)>t| S_0=s'),\\
     &= 1+ \sum_{s'} d_{\pi}(s')\sum_{t \ge 1} P_{\pi}(\tau^+(s)>t| S_0=s') -\sum_{s' \neq s} d_{\pi}(s') \sum_{t \ge 0} P_{\pi}(\tau^+(s)>t| S_0=s'),\\
     &= 1+\sum_{s'} d_{\pi}(s')\sum_{t \ge 1} P_{\pi}(\tau^+(s)>t,S_1 \neq s| S_0=s') -\sum_{s' \neq s} d_{\pi}(s') \sum_{t \ge 0} P_{\pi}(\tau^+(s)>t| S_0=s'),\\
     &= 1+ \sum_{s' \neq s} d_{\pi}(s') \sum_{t \ge 0} P_{\pi}(\tau^+(s)>t| S_0=s')-\sum_{s' \neq s} d_{\pi}(s') \sum_{t \ge 0} P_{\pi}(\tau^+(s)>t| S_0=s'),\\
     &=1.
\end{align*}
Here, we use $\tau^+_s(s)$ to represent the visitation time to state $s$ after step $t=0$ starting from state $s$. The fourth line comes from that if $S_0$ follows the stationary distribution, then $S_1$ follows it as well. Meanwhile, $\tau^+(s)>1$ implies that $S_1 \neq s$. Finally, in the last line, the sums of these two series can cancel each other since they are all finite due to states' positive recurrences. Now the proof is done.
\end{proof}

\section{Proof of the Extended Policy Gradient Theorem}


Then, let us study its existence under limits and its correctness. But before all, let us present the ergodic theorem in Norris (1998) and a limit of product lemma for future proof.

\begin{theorem}[Theorem 1.0.2 Ergodic theorem, Norris (1998)]
Let $\mathcal{M}$ be an irreducible and positive recurrent Markov decision process for all policies. Then, for each state $s$,
\begin{align*}
    \mathbb{P}(\frac{\sum_{t=0}^{T-1} \mathds{1}[S_t=s]}{T} \to \frac{1}{\mathbb{E}[\tau_s^+(s)]} \text{ as } T \to \infty) =1.
\end{align*}
\end{theorem}

\begin{lemma} \label{lemma: product_sequences}
Let $(a_n)$ and $(b_n)$ be two convergence sequences with limits $a$ and $b$. If $(a_n)$ is bounded for all $n \ge 1$ and the limit $b$ is also bounded, then $$ab=\lim_{n_1\to\infty} a_{n_1} \lim_{n_2 \to \infty} b_{n_2}=\lim_{n\to\infty} a_n b_n.$$
\end{lemma}

Then, we show that our correction terms exist and are finite for all positive recurrent states. The proof is straightforward, following the ergodic theorem.

\begin{lemma}
For any positive recurrent state $s$, the limit of averaging correction $c(s)$ exists and is bounded by the expected recurrence time:
\begin{equation*}
    c(s) = \mathbb{E}[\tau^+_s(s)] d_{\pi,\gamma}(s)\le \mathbb{E}[\tau^+_s(s)] < \infty.
\end{equation*}
\end{lemma}
\begin{proof}
First, the ergodic theorem tells that, almost surely, the following equality holds:
\begin{align*}
    \lim_{T \to \infty} (1-\gamma) T  \frac{ \sum_{t = 0}^{T-1} \gamma^{t} \mathds{1}[S_{t} = s] }{ \sum_{k = 0}^{T-1} \mathds{1}[S_{k} = s] }&=\lim_{T \to \infty} \frac{ T }{ \sum_{k = 0}^{T-1} \mathds{1}[S_{k} = s] } \cdot \lim_{T\to\infty}(1-\gamma)\sum_{t = 0}^{T-1} \gamma^{t} \mathds{1}[S_{t} = s] \\
    &= \E[\tau^+_s(s)] \cdot \sum_{t=0}^{\infty} (1-\gamma)\gamma^t \mathds{1}[S_t=s].
\end{align*}
The first equality holds with two bounded sequences according to Lemma \ref{lemma: product_sequences}.

By dominated convergence theorem, this convergence can be brought inside the expectation:
\begin{align*}
    c(s) & = \E \left[ \lim_{T \to \infty} (1-\gamma) T  \frac{ \sum_{t = 0}^{T-1} \gamma^{t} \mathds{1}[S_{t} = s] }{ \sum_{k = 0}^{T-1} \mathds{1}[S_{k} = s] } \right],\\
    & = \E[ \E[\tau^+_s(s)] \cdot \sum_{t=0}^{\infty} (1-\gamma)\gamma^t \mathds{1}[S_t=s]],\\
    &= \E[\tau^+_s(s)] \cdot \sum_{t=0}^{\infty} (1-\gamma)\gamma^t P_{\pi}(S_t=s),\\
    &= \E[\tau^+_s(s)] \cdot d_{\pi,\gamma}(s)\\
    &\le \E[\tau^+_s(s)].
\end{align*}
\end{proof}

Next, we present our extended policy gradient theorem and show that our correction term equals the distribution ratio, $\frac{d_{\pi,\gamma}(s)}{d_{\pi}(s)}$ and thus, this term successfully corrects the state distribution mismatch.

\begin{theorem}[Extended Policy Gradient Theorem]
Given an irreducible Markov decision process with positive recurrent states, the gradient of the discounted objective can be expressed as:
\begin{align}
\nabla J(\pi)&= \mathbb{E}_{s\sim d_{\pi,\gamma}(s)}[\sum_{a \in\mathcal{A}}\nabla \pi(a|s)q_{\pi,\gamma}(s,a)] \\
&= \mathbb{E}_{s \sim d_{\pi}(s)}[c(s)\sum_{a \in\mathcal{A}}\nabla \pi(a|s)q_{\pi,\gamma}(s,a)] 
\end{align}
\end{theorem}

\begin{proof}
The expected recurrence time can be changed to the reverse of the stationary state distribution according to the multiple analytical forms of the stationary distribution. Then, our correction equals the state distribution ratio, and the proof is done.
\begin{align*}
    c(s) 
    &= \E[\tau^+_s(s)] \cdot d_{\pi,\gamma},\\
    &= \frac{d_{\pi,\gamma}(s)}{d_{\pi}(s)}.
\end{align*}
\end{proof}

Moreover, the correctness of the existing correction follows from the extended policy gradient theorem, shown in the following corollary. Thus, our paper first models and proves that the existing solution multiplied by the discount factor's powers fixes the bias.

\begin{corollary}
Given an irreducible Markov decision process with positive recurrent states, the gradient of the discounted objective can be written as:
\begin{align*}
    &\nabla J(\pi) = \mathbb{E}_{s\sim d_{\pi,\gamma}(s)}[\sum_{a \in\mathcal{A}}\nabla \pi(a|s)q_{\pi,\gamma}(s,a)],\\
    &= \lim_{T \to \infty} \E_{s\sim d_{\pi}(s),K_T(s)}\left[ T(1-\gamma)\gamma^{K_T(s)} \sum_{a\in\mathcal{A}}\nabla \pi(a|s) q_{\pi,\gamma}(s,a)\right].
\end{align*}
\end{corollary}
\begin{proof}
\begin{align*}
    &\E_{s\sim d_{\pi}(s),K_T(s)}\left[ \lim_{T \to \infty}  T(1-\gamma)\gamma^{K_T(s)} \sum_{a\in\mathcal{A}}\nabla \pi(a|s) q_{\pi,\gamma}(s,a)\right] \\
    & = \E_{s\sim d_{\pi}(s)}\left[ c(s) \sum_{a\in\mathcal{A}}\nabla \pi(a|s) q_{\pi,\gamma}(s,a)\right],\\
    &=\nabla J(\pi).
\end{align*}
\end{proof}



\begin{table}[t]
    \centering
    \small
    \begin{tabular}{|c|c|c|c|c|c|c|c|c|}
        \hline
         Hyperparameters & \multicolumn{4}{c|}{Biased} & \multicolumn{4}{c|}{Existing Correction}  \\
        \hline
         & $\gamma =0.3$ & $\gamma =0.5$& $\gamma =0.7$ & $\gamma =0.9$& $\gamma =0.3$ & $\gamma =0.5$& $\gamma =0.7$ & $\gamma =0.9$ \\
         \hline
         Learning rate &0.95&0.95&0.95&0.95&0.95&0.95&0.95&0.95\\
         \hline
         Batch size & 8&8&8&8&8&8&8&1\\
         \hline
    \end{tabular}
    \begin{tabular}{|c|c|c|c|c|}
        \hline
         Hyperparameters & \multicolumn{4}{c|}{Our Correction}  \\
        \hline
         & $\gamma =0.3$ & $\gamma =0.5$& $\gamma =0.7$ & $\gamma =0.9$ \\
         \hline
         Learning rate &0.95&0.95&0.95&0.5\\
         \hline
         Batch size & 1&1&1&1\\
         \hline
         Correction learning rate & 0.008&0.01&0.005&0.005\\
         \hline
    \end{tabular}
    \caption{Hyperparameters for three agents on the counterexample.}
    \label{tab:hyper_counter}
\end{table}

\section{Proof of Error Bound on Approximated Correction}
\begin{proposition}
Consider an irreducible and positive recurrent MDP under policy $\pi$ with a finite state space $\mathcal{S}$. Given a data buffer $\mathcal{D}$ consisting of $k$ trajectories under a policy $\pi$, each with length $T$, if $$k \ge \frac{2}{\epsilon^2}\log\frac{|\mathcal{S}|}{\delta}$$ and $$T \ge \cfrac{\log \frac{\epsilon}{2}}{\log \gamma},$$ then with probability at least $1-\delta$, the error is smaller than $\epsilon$, that is
\begin{align*}
    \max _{s\in\mathcal{S}}|\hat{d}_{\mathcal{D}}(s) c_{\mathcal{D}}(s) - d_{\pi,\gamma}(s)| \le \epsilon.
\end{align*}
\end{proposition}
\begin{proof}
Define a random variable $X_j$ dependent on the $j$-th trajectory in the data buffer as 
$$X_j = (1-\gamma)\sum_{t=0}^{T-1} \gamma^t \mathds{1}[S_{t,j} =s].$$

So we can rewrite the approximated correction as
\begin{align*}
    c_{\mathcal{D}}(s)  &=\frac{|\mathcal{D}|}{\sum_{i=1}^{|\mathcal{D}|} \mathds{1}[S_i=s]} \frac{1}{k}(1-\gamma)\sum_{j=1}^k \sum_{t=0}^{T-1}\gamma^t \mathds{1}[S_{t,j} =s] \\
    & = \frac{|\mathcal{D}|}{\sum_{i=1}^{|\mathcal{D}|} \mathds{1}[S_i=s]} \frac{1}{k}\sum_{j=1}^k X_j\\
     & = \frac{1}{\hat{d}_{\mathcal{D}}(s)} \frac{1}{k}\sum_{j=1}^k X_j.\\
\end{align*}

Thus for each state $s$, the error term can be bounded as
\begin{align*}
    |\hat{d}_{\mathcal{D}}(s) c_{\mathcal{D}}(s)   - d_{\pi,\gamma}(s)| 
    &= |\hat{d}_{\mathcal{D}}(s) \frac{1}{\hat{d}_{\mathcal{D}}(s)} \frac{1}{k}\sum_{j=1}^k X_j - d_{\pi,\gamma}(s)|\\
    & = |\frac{1}{k}\sum_{j=1}^k X_j - d_{\pi,\gamma}(s)|\\
    & \le |\frac{1}{k}\sum_{j=1}^k X_j - (1-\gamma) \sum_{t=0}^{T-1} \gamma^t \mathbb{P}_{\pi}(S_t=s)| +| (1-\gamma) \sum_{t=0}^{T-1} \gamma^t \mathbb{P}_{\pi}(S_t=s) -d_{\pi,\gamma}(s)|\\
    & = |\frac{1}{k}\sum_{j=1}^k X_j - (1-\gamma) \sum_{t=0}^{T-1} \gamma^t \mathbb{P}_{\pi}(S_t=s)| +| (1-\gamma) \sum_{t\ge T} \gamma^t \mathbb{P}_{\pi}(S_t=s)|\\
    & \le |\frac{1}{k}\sum_{j=1}^k X_j - (1-\gamma) \sum_{t=0}^{T-1} \gamma^t \mathbb{P}_{\pi}(S_t=s)| +(1-\gamma) \sum_{t\ge T} \gamma^t \\
    & \le |\frac{1}{k}\sum_{j=1}^k X_j - (1-\gamma) \sum_{t=0}^{T-1} \gamma^t \mathbb{P}_{\pi}(S_t=s)| +\gamma^T. \\
\end{align*}
When $T \ge \cfrac{\log \frac{\epsilon}{2}}{\log \gamma}$, we have $\gamma^T \le \frac{\epsilon}{2}$.

 Notice that each random variable $X_j$ depends on the corresponding trajectory; thus, $X_j,\:j=1,\cdots,k$ are independent. Meanwhile, their expectations are the same and for all $j=1,\cdots,k$, the expectation equals
 \begin{align*}
     \mathbb{E}[X_j] = (1-\gamma) \sum_{t=0}^{T-1} \gamma^t \mathbb{P}_{\pi}(S_t=s).
 \end{align*}

 Thus, we can bound the other part with Hoeffding's inequality. For each state $s$, with probability at most $\frac{\delta}{|\mathcal{S}|}$, when $k \ge \frac{2}{\epsilon^2}\log\frac{|\mathcal{S}|}{\delta}$, we have
 \begin{align*}
     |\frac{1}{k}\sum_{j=1}^k X_j - (1-\gamma) \sum_{t=0}^{T-1} \gamma^t \mathbb{P}_{\pi}(S_t=s)| \le \frac{\epsilon}{2}.
 \end{align*}
\end{proof}

The propositions says if we want to bound the distribution error by $\epsilon$, we need $k \ge \frac{2}{\epsilon^2}\log\frac{|\mathcal{S}|}{\delta}$ and $T \ge \cfrac{\log \frac{\epsilon}{2}}{\log \gamma}$. In a buffer, $k$ is the number of trajectories, $|\mathcal{D}|$ is the number of samples and $T$ is the trajectory length. So $|\mathcal{D}|=kT$. Here, if we want to analyze the relationship between the error $\epsilon$, and $k$ and $T$, we can use big-O to cover all other terms. Then, the error $\epsilon$ depends on $\frac{1}{\sqrt{k}}$ and $\gamma^T$. Let's write $k = \frac{|\mathcal{D}|}{T}$. So we have the error is of order $O(\max\{\sqrt{\frac{T}{|\mathcal{D}|}},\gamma^T\})$. As $k$ and $T$ gets larger, $\frac{1}{\sqrt{k}}$ and $\gamma^T$ get smaller and so is the error.

\section{Counterexample}
\begin{figure}[!htb]
\hspace*{-5.3mm}
\minipage{0.32\textwidth}
  \includegraphics[width=1.12\linewidth]{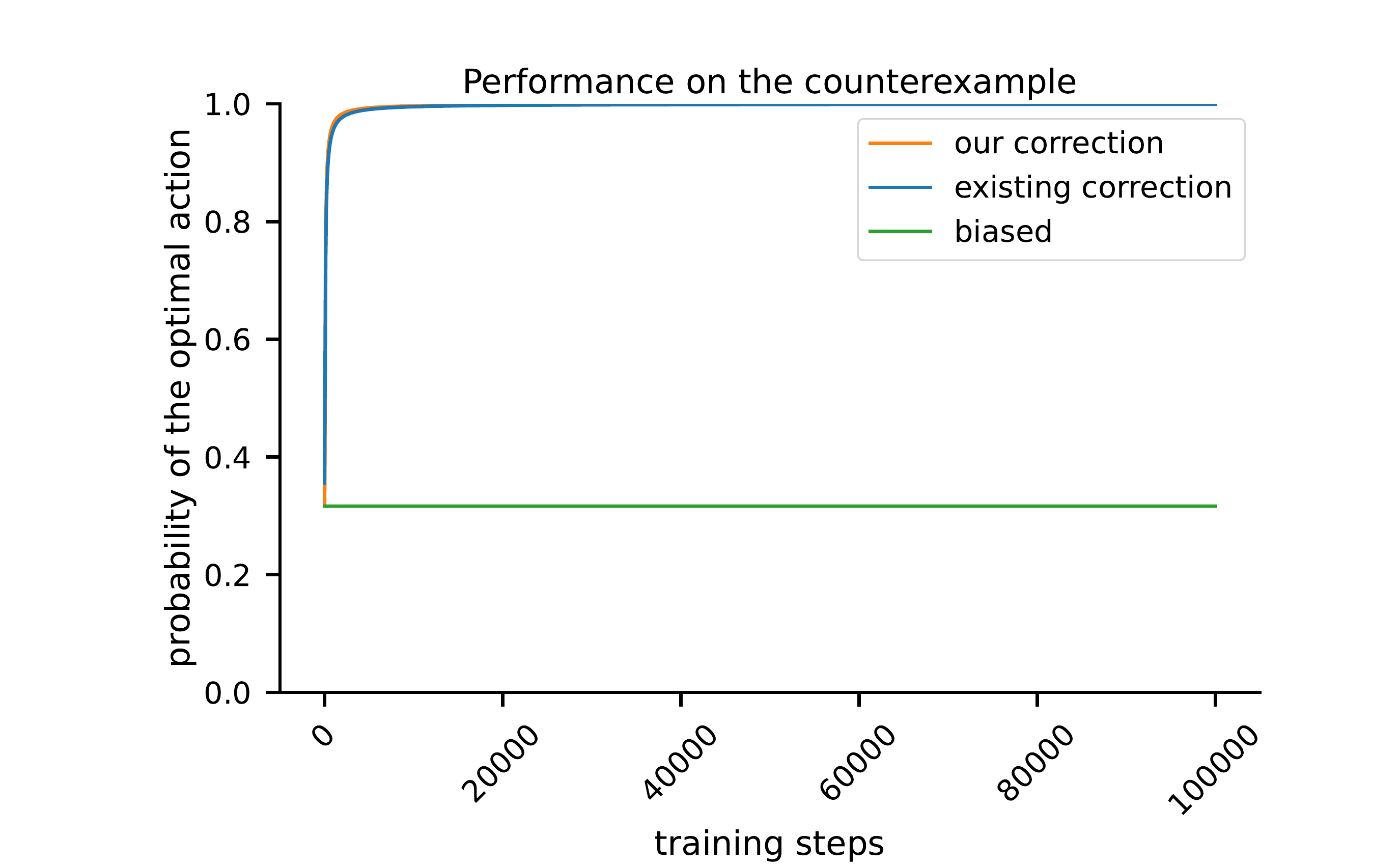}
  \hspace*{-2mm}
\endminipage\hfill
\minipage{0.32\textwidth}
  \includegraphics[width=1.12\linewidth]{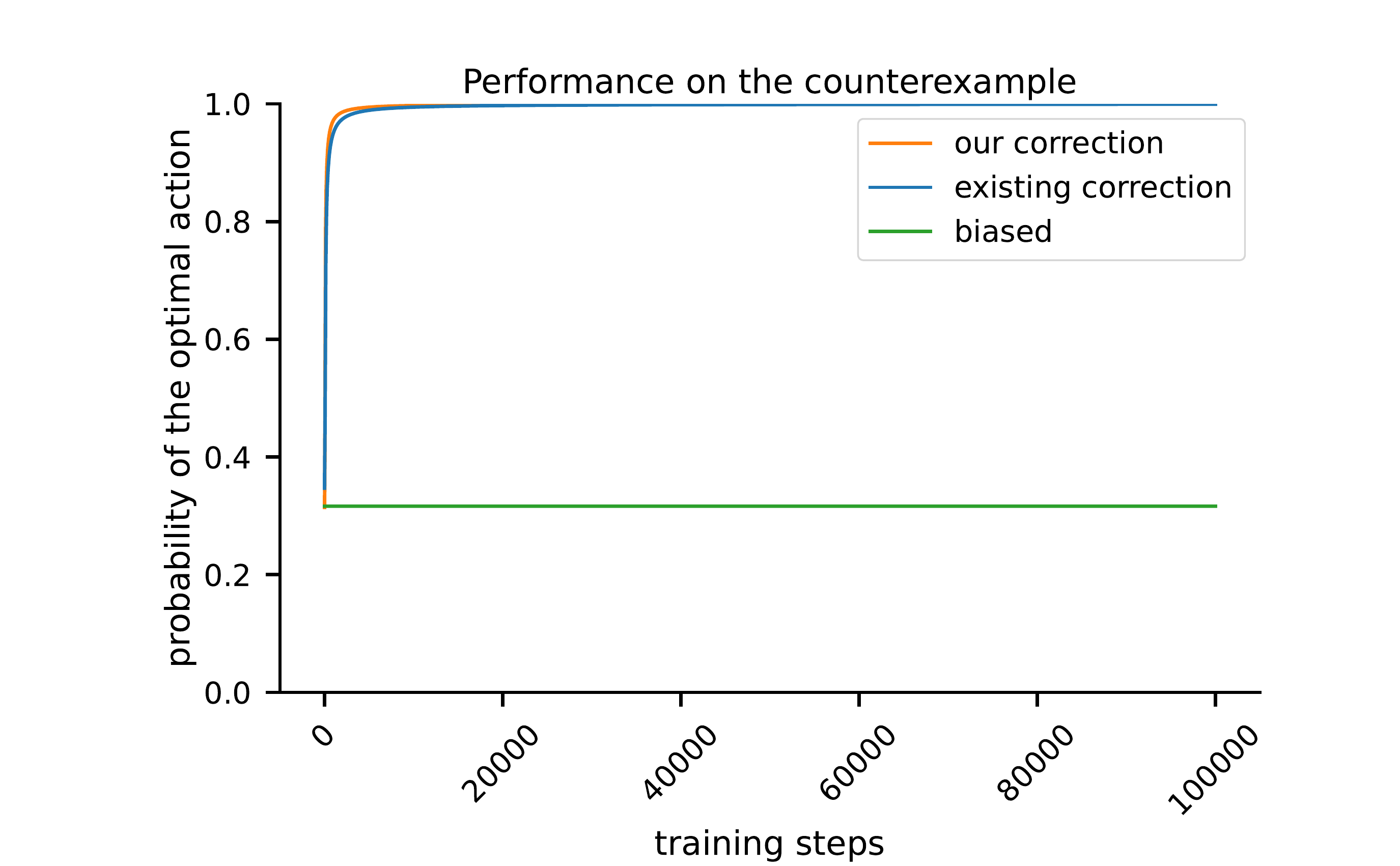}
  \hspace*{-2mm}
\endminipage\hfill
\minipage{0.32\textwidth}%
  \includegraphics[width=1.12\linewidth]{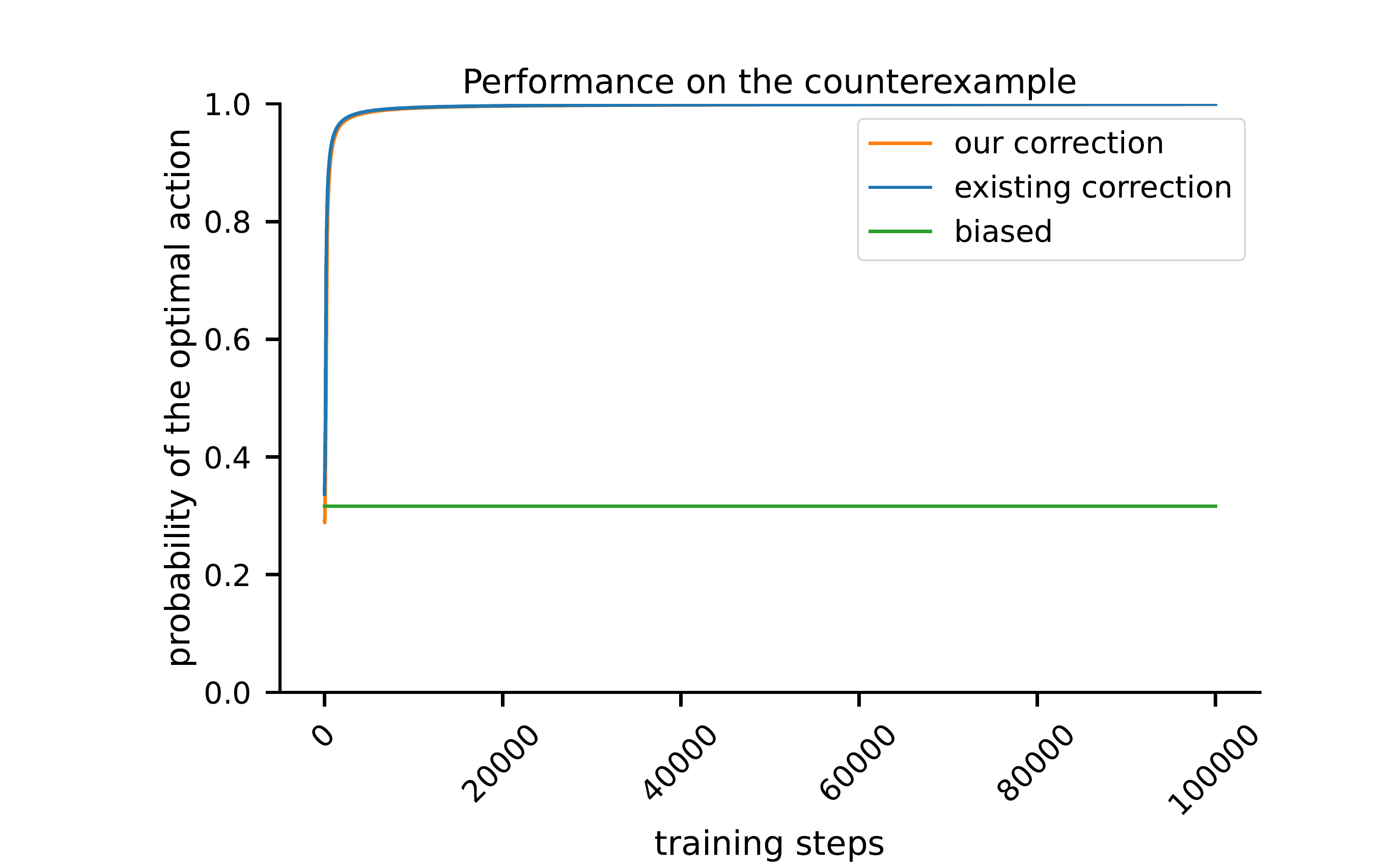}
\endminipage
\vspace*{-3mm}
\caption{These figures present the learnt probability
of choosing the optimal action on the counterexample when the agents use the original
biased gradient and gradients with corrections. The above figures use discount factors $0.3$, $0.5$, and $0.7$.}
\label{fig:counterexample_gammas}
\end{figure}

For different choices of discount factors, the agents with the original biased gradients shown in green lines keep choosing the optimal action with the initialized probability, not learning to choose the optimal action all the time. It tells us that corrections of state distributions are needed for a wide range of discount factors.

All these three agents utilize the true Q-values to compute gradients. The hyperparameters are tuned for each discount factor, shown in Table \ref{tab:hyper_counter}. Even though our correction introduces a new hyperparameter, a learning rate for the correction network, the computations when tuning hyperparameters for all three agents are the same. Each agent samples $500$ possible combinations of hyperparameters.

\begin{figure}
    \minipage{0.49\textwidth}
    \hspace*{-4mm}
    \includegraphics[width=1.12\linewidth]{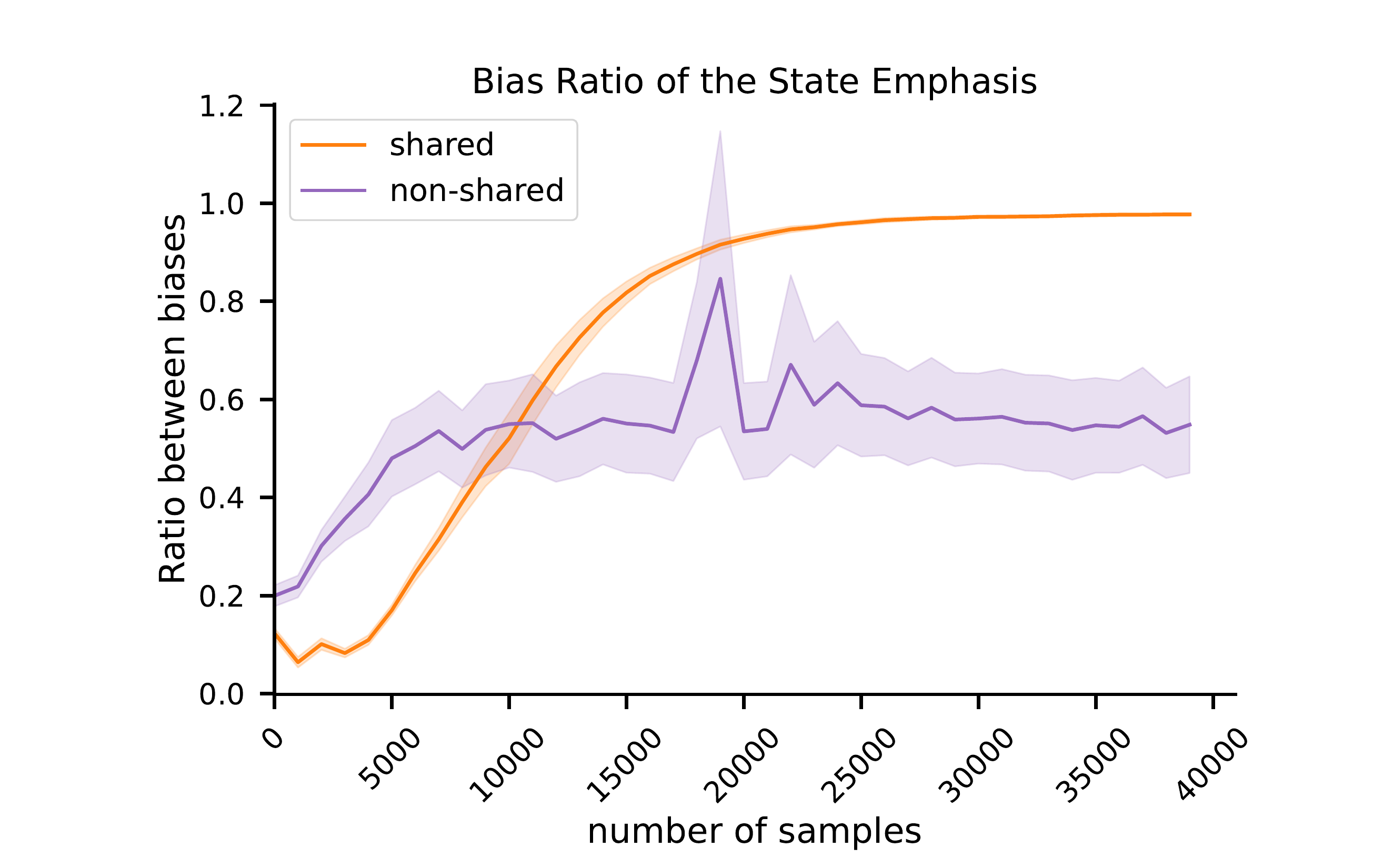}
    \caption{The figure shows the bias ratios of state emphasis with our corrections over the bias from the original biased gradient. Less than one ratio proves that our approximated correction successfully reduces the bias from the discount-factor mismatch. We test two architectures to determine whether the correction network shares the first two layers with the value network or not. }
    \label{fig:bias_ratio_shared}
    \endminipage\hfill
    \minipage{0.49\textwidth}
    \hspace*{-4mm}
    \includegraphics[width=1.12\linewidth]{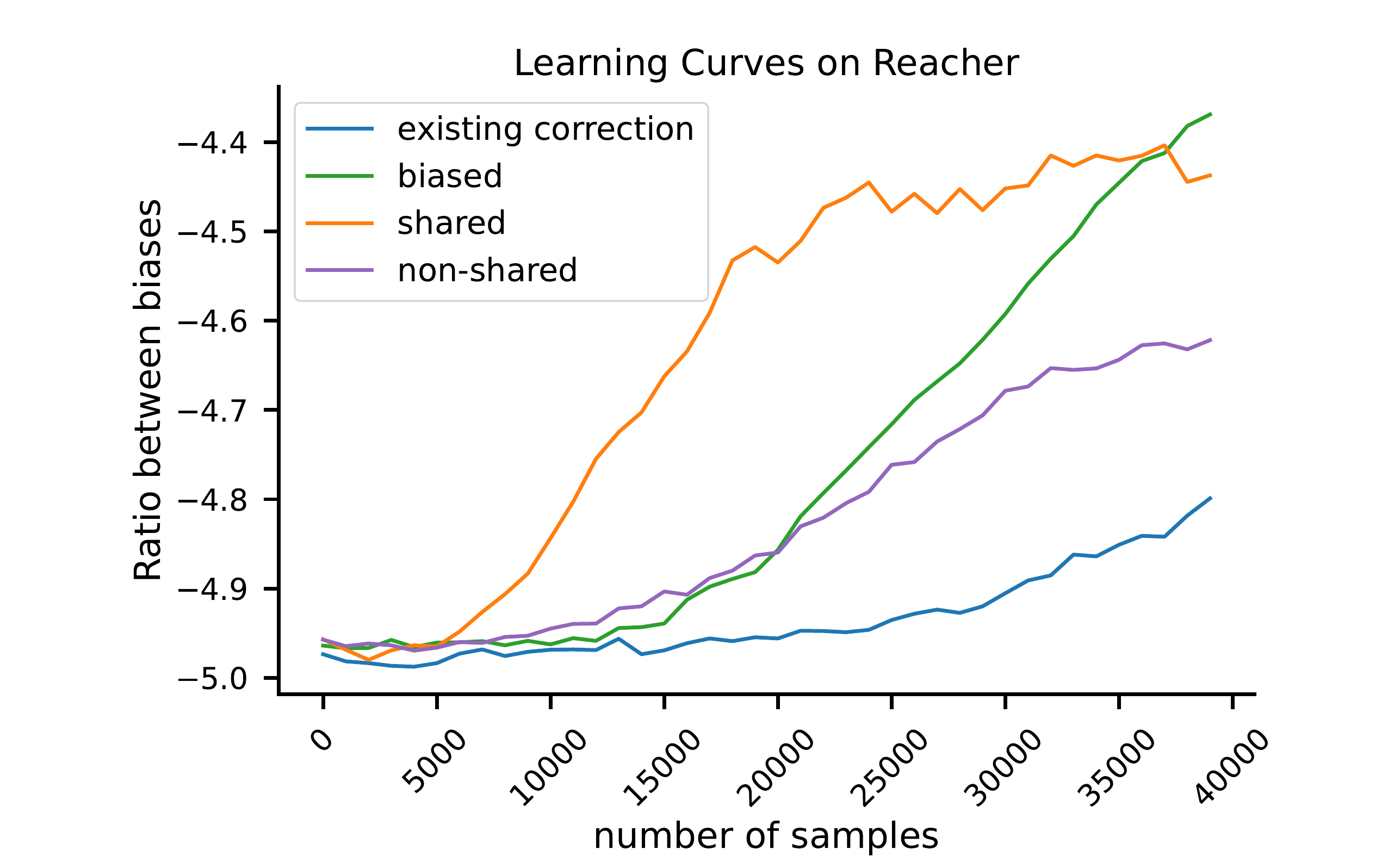}
    \caption{The figure shows the averaged returns for three agents during learning, over $10$ random seeds. Our correction with the shared correction network has the fastest learning speed, and its fast learning result matches the early lower bias result.}
    \label{fig:reacher_rets}
    \hspace*{-2mm}
    \endminipage\hfill
\end{figure}

\section{Bias Reduction Analysis}
For the result in Fig.\ \ref{fig:bias_ratio}, we train a batch actor-critic agent with our correction under one random seed. At various checkpoints, we collect ten buffers of data and compute the state weightings used by three agents under the current policy. The x-axis shows the number of training samples. 

Then, we also numerically study the bias-reduction performance; the performance is measured by the ratio between our approximation error and the state distribution mismatch error, both averaged over the buffer $\mathcal{D}$, that is
\begin{align*}
    \cfrac{\sum_{i=1}^{|\mathcal{D}|}|\hat{c}_{\mathcal{D}}(s_i)\hat{d}_{\mathcal{D}}(s_i)- d_{\pi,\gamma}(s_i)|}{\sum_{i=1}^{|\mathcal{D}|}|\hat{d}_{\mathcal{D}}(s_i)- d_{\pi,\gamma}(s_i)|},
\end{align*}
where $d_{\pi,\gamma}$ is the correct discounted stationary state distribution, $\hat{d}_{\mathcal{D}}$ is the sampling distribution from the buffer and $\hat{c}_{\mathcal{D}}$ is our approximated correction. A ratio smaller than one means that the approximation error of our estimator is less than the state distribution mismatch error, and thus the estimator successfully reduces the bias.

Furthermore, we test different neural networks' architectures for our algorithm according to the bias reduction performance, and whether the correction network shares the first two layers with the value network. The hyperparameters are tuned for each architecture as shown in Table \ref{tab:reacher_hyper}. Our algorithm introduces two new hyperparameters: a learning rate for the correction network and a constant scaling up the power term $\gamma^t$ to avoid tiny values. 

As shown in Figure \ref{fig:bias_ratio_shared}, both architectures give a ratio smaller than one. The result tells that our algorithm successfully diminishes the bias introduced by the discount-factor mismatch. Meanwhile, the shared network provides smaller biases than the non-shared one before $10k$ steps and may show learning advantage earlier. Thus, we leverage the shared network for all our algorithms. 

\begin{table}[h]
    \centering
    \small
    \begin{tabular}{|c|c|c|c|c|c|c|}
    \hline
     Algorithm  & Batch size & Learning rate & Critic learning rate & Critic loss weight & scale & $\#$ hidden unit for critic \\
     \hline
    Shared& 25 & 0.0042 & 0.0005&18.94 &143&64\\
    \hline
    Non-shared & 45 & 0.0047 & 0.0003 & 8.9 & 133 &32 \\
    \hline
    \end{tabular}
    \begin{tabular}{|c|c|c|}
    \hline
     Algorithm  &  $\#$ hidden unit for policy & gamma\\
     \hline
    Shared&8&0.99\\
    \hline
    Non-shared &  32 & 0.8\\
    \hline
    \end{tabular}
    \caption{Hyperparameters for batch actor-critic with our correction on a discrete Reacher.}
    \label{tab:reacher_hyper}
\end{table}

\section{Batch Actor-Critic}
For all our experiments, the actor-critic was parameterized using two-layer neural networks. The correction network is based on the hidden layers of the critic and adds one more layer, leading to the distribution correction term. 

Even though our correction introduces two new hyperparameters: a learning rate for the correction network and a constant scaling up of the power term $\gamma^t$ to avoid tiny values, the computations when tuning hyperparameters for all three agents are the same. Each agent samples $300$ possible combinations of hyperparameters using random search \cite{bergstra2012random}.

The key hyperparameters for the original algorithm and the algorithm with the existing correction include the learning rates for both policy and value networks, the number of hidden units for both policy and value networks, buffer size, the coefficient for critic loss and the discount factor. The hyperparameter values are listed in Table \ref{tab:BAC_hyper}. 

\begin{table}[h]
    \centering
    \small
    \begin{tabular}{|c|c|c|c|c|c|c|}
    \hline
     Algorithm  & Batch size & Learning rate & Critic learning rate & Critic loss weight & scale & $\#$ hidden unit for critic \\
     \hline
    biased& 64 & 0.0008 & 0.0008&8.8 &None&128\\
    \hline
    existing correction & 64 & 0.0008 & 0.0008&8.8 &None&128\\
    \hline
    our correction & 128 & 0.0003 & 0.009 &9.6 & 29 &64 \\
    \hline
    \end{tabular}
    \begin{tabular}{|c|c|c|}
    \hline
     Algorithm  &  $\#$ hidden unit for policy & gamma\\
     \hline
    biased&32&0.995\\
    \hline
    existing correction &  32 & 0.995\\
    \hline
    our correction &  32 & 0.995\\
    \hline
    \end{tabular}
    \caption{Tuned hyperparameters values for BAC and modified BACs on CartPole and are used for other openAI gym tasks. The value is left None if a hyperparameter is not tuned for that algorithm. }
    \label{tab:BAC_hyper}
\end{table}


\section{PPO}
The original PPO algorithm followed the implementation by OpenAI spinningup. The hyperparameters are tuned for each algorithm on each task. We list the values of hyperparameters in Table \ref{tab:PPO_hyper}. Each agent samples $900$ possible combinations of hyperparameters using random search \cite{bergstra2012random} on MuJoCo tasks and $250$ possible combinations on Mountain Car and Point Mass.

\subsection{Hyperparameters}
\begin{table}[h]
    \centering
    \small
    \textbf{HalfCheetah}
    \begin{tabular}{|c|c|c|c|c|c|c|c|}
    \hline
     Algorithm  &  Learning rate & Target KL &Critic learning rate & Critic loss weight & scale & critic hidden units  & gamma\\
     \hline
    biased& 0.0009 & 0.01& 0.0047&None&None&64&0.99\\
    \hline
    existing correction & 0.0009 & 0.01&0.0011& None &None&256&0.99\\
    \hline
    our correction &  0.0009 & 0.01&0.0026 &4.0 & 49 &64&0.99 \\
    \hline
    \end{tabular}
    \textbf{Ant}
    \begin{tabular}{|c|c|c|c|c|c|c|c|}
    \hline
     Algorithm  &  Learning rate & Target KL &Critic learning rate & Critic loss weight & scale & critic hidden units  & gamma\\
     \hline
    biased& 0.0009 & 0.02& 0.0007&None&None&64&0.97\\
    \hline
    existing correction & 0.0003 & 0.02&0.0008& None &None&128&0.99\\
    \hline
    our correction & 0.0003 & 0.02&0.0008& 113 &0.66&128&0.99 \\
    \hline
    \end{tabular}
    \textbf{Walker}
    \begin{tabular}{|c|c|c|c|c|c|c|c|}
    \hline
     Algorithm  &  Learning rate & Target KL &Critic learning rate & Critic loss weight & scale & critic hidden units  & gamma\\
     \hline
    biased& 0.0009 & 0.01& 0.0023&None&None&64&0.99\\
    \hline
    existing correction & 0.0003 & 0.11&0.0042& None &None&128&0.995\\
    \hline
    our correction & 0.0003 & 0.29&0.0004& 34 &2.82&256&0.995 \\
    \hline
    \end{tabular}
    \textbf{Swimmer}
    \begin{tabular}{|c|c|c|c|c|c|c|c|}
    \hline
     Algorithm  &  Learning rate & Target KL &Critic learning rate & Critic loss weight & scale & critic hidden units  & gamma\\
     \hline
    biased& 0.003 & 0.01& 0.0027&None&None&256&0.995\\
    \hline
    existing correction & 0.0003 & 0.11&0.0042& None &None&128&0.995\\
    \hline
    our correction & 0.003 & 0.01& 0.0027&None&None&256&0.995 \\
    \hline
    \end{tabular}
    \textbf{Mountain Car Continuous}
    \begin{tabular}{|c|c|c|c|c|c|c|c|}
    \hline
     Algorithm  &  Learning rate & Target KL &Critic learning rate & Critic loss weight & scale & critic hidden units  & gamma\\
     \hline
    biased& 0.0003 & 0.08& 0.0045&None&None&256&0.995\\
    \hline
    existing correction & 0.0003 & 0.22&0.0047& None &None&256&0.995\\
    \hline
    our correction & 0.0009 & 0.27& 0.0044&1.64&104&128&0.995 \\
    \hline
    \end{tabular}
    \textbf{Point Mass}
    \begin{tabular}{|c|c|c|c|c|c|c|c|}
    \hline
     Algorithm  &  Learning rate & Target KL &Critic learning rate & Critic loss weight & scale & critic hidden units  & gamma\\
     \hline
    biased& 0.0003 & 0.08& 0.0041&None&None&64&0.99\\
    \hline
    existing correction & 0.0009 & 0.2&0.0007& None &None&128&0.99\\
    \hline
    our correction & 0.0009 & 0.06& 0.0028&3.28&12&256&0.995 \\
    \hline
    \end{tabular}
    \caption{Tuned hyperparameters values for PPOs.}
    \label{tab:PPO_hyper}
\end{table}

Our correction can give learning advantages over the other two baselines in several environments as shown in Figure \ref{fig:mujoco_multiple} if it adopts results from three combinations of hyperparameters. All hyperparameters are tuned on Hopper and Ant. and generalize well. Our correction shows unbeatable performances on untuned tasks: Walker2d, Inverted Double Pendulum, Swimmer and Reacher. 


\subsection{dm\_control Reacher using Sparse Rewards}
The \emph{reacher} task aims to move the fingertip of a planar arm with two degrees of freedom (DoF) to a random spherical target on a 2D plane. 
It has two sub-tasks: easy and hard. 
They differ in the sizes of the target and the fingertip of the arm. 
The observation includes the position of the fingertip, the speed of the fingertip, and the vector from the fingertip to the target. 
The action space is the torques applied to the two joints, scaled to the range $-1$ to $1$. 
We modified the Deepmind Control Suite reacher task \cite{tassa2018deepmind} such that the episode terminates upon reaching the goal state.
The reward function is modified to give $-1$ for each step to encourage shorter episodes. 
After each \textit{timeout}, we reset the agent by moving the fingertip to a random location on the plane while keeping the target unchanged. 
This process continues until the fingertip reaches the target within the target size. 
Once the target is reached, the current episode terminates.
At episode terminations, we reset the agent and randomly generate a new target for the next episode. 

We tuned the hyper-parameters listed in table \ref{tab:DMeasy} and \ref{tab:DMhard} for all three variants. We tested $200$ combinations of hyper-parameters each using random search \cite{bergstra2012random}. 

\begin{table}[h]
    \centering
    \small
    \begin{tabular}{|c|c|c|c|c|c|c|}
    \hline
     Algorithm  & Critic loss weight & Scale & Target KL & Critic learning rate & Gamma & $\#$ hidden units (actor, critic) \\
     \hline
    biased & 8.53 & 135 & 0.258 & 0.0005 & 0.99 & 64, 64\\
    \hline
    existing correction & 7.09  & 83 & 0.197 & 0.004 & 0.99 & 256, 256 \\
    \hline
    our correction & 8.04 & 67 & 0.42 & 0.0017 & 0.995 & 256, 256 \\
    \hline
    \end{tabular}
    \caption{Choice of hyper-parameters for dm\_reacher\_easy}
    \label{tab:DMeasy}
\end{table}

\begin{table}[h]
    \centering
    \small
    \begin{tabular}{|c|c|c|c|c|c|c|}
    \hline
     Algorithm  & Critic loss weight & Scale & Target KL & Critic learning rate & Gamma & $\#$ hidden units (actor, critic) \\
     \hline
    biased & 5.05 & 139 & 0.443 & 0.0015 & 0.9 & 64, 256\\
    \hline
    existing correction & 1.32 & 43 & 0.328 & 0.0021 & 0.99 & 256, 64\\
    \hline
    our correction & 1.12 & 93 & 0.275 & 0.0045 & 0.995 & 64, 256 \\
    \hline
    \end{tabular}
    \caption{Choice of hyper-parameters for dm\_reacher\_hard}
    \label{tab:DMhard}
\end{table}

\begin{figure}[!ht]
    \minipage{0.49\textwidth}
    \hspace*{-4mm}
    \includegraphics[width=\linewidth]{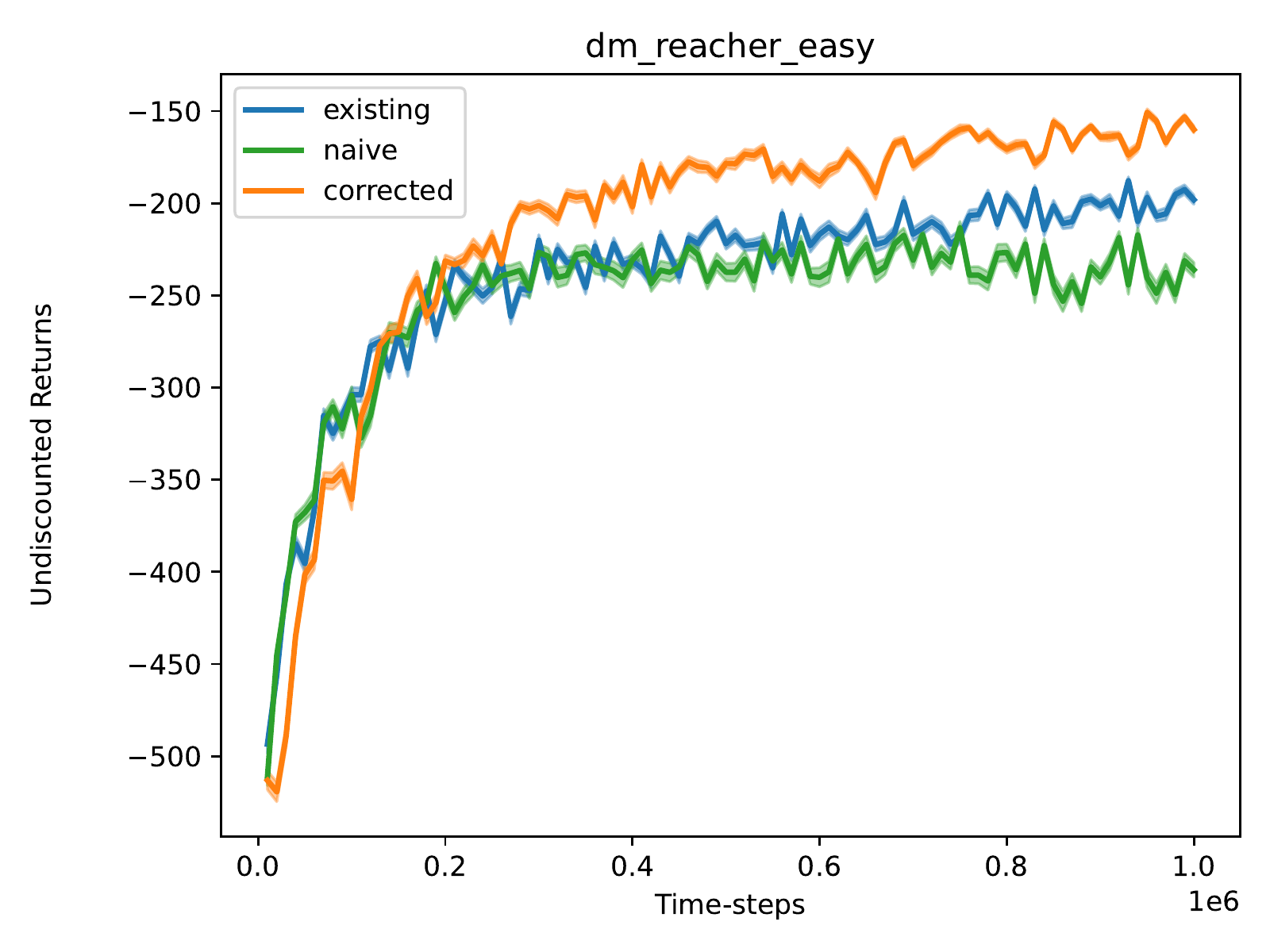}
    \caption{}
    \label{fig:dm_reacher_easy}
    \endminipage\hfill
    \minipage{0.49\textwidth}
    \hspace*{-4mm}
    \includegraphics[width=\linewidth]{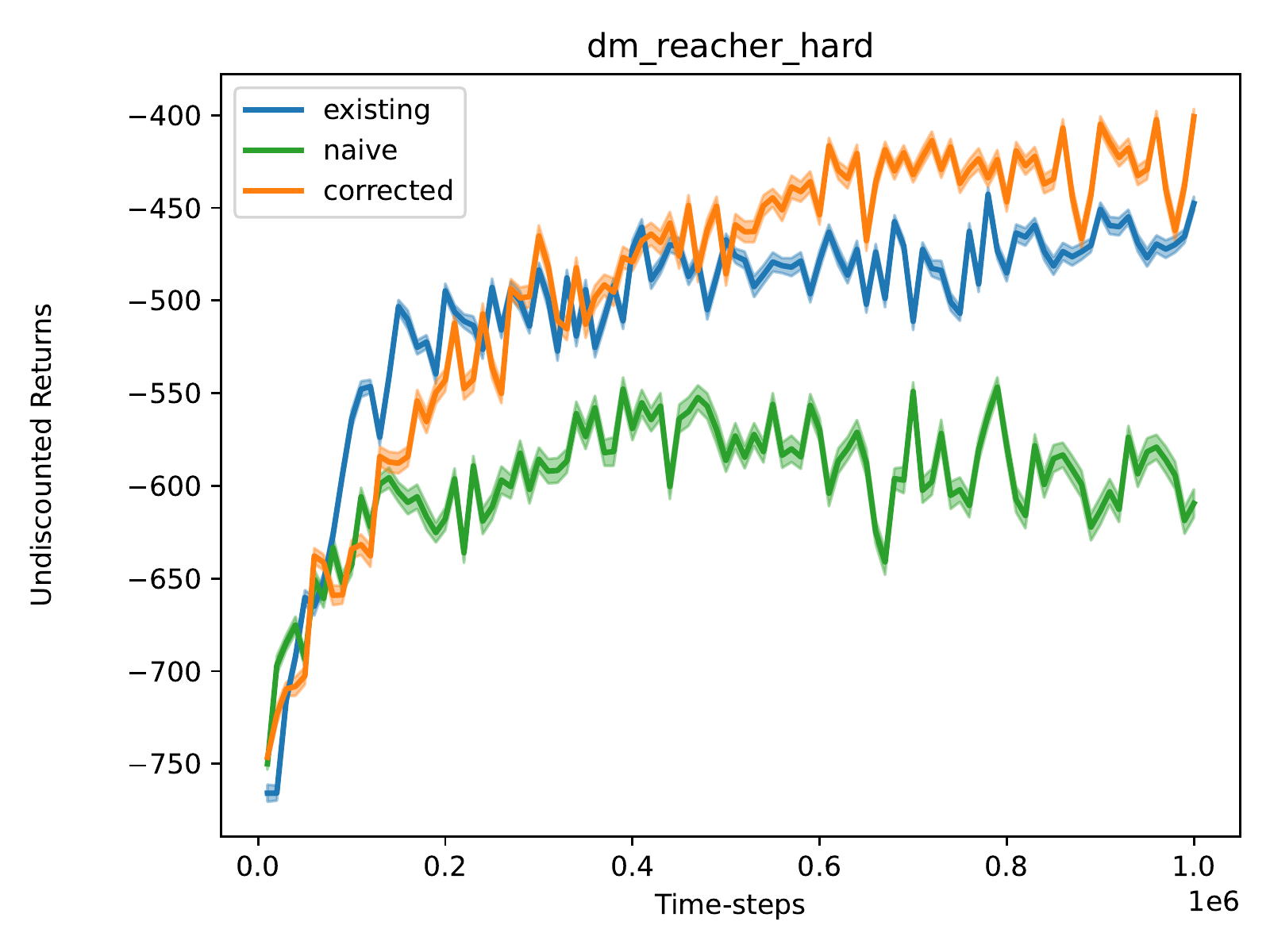}
    \caption{}
    \label{fig:dm_reacher_hard}
    \hspace*{-2mm}
    \caption{Performance of modified PPO and two baselines on modified dm\_control reacher tasks averaged over 30 runs}
    \endminipage\hfill
\end{figure}

\subsection{UR5-Reacher Experiments with PPO}
\begin{figure*}[!ht]
    \centering
    \includegraphics[width=0.5\textwidth]{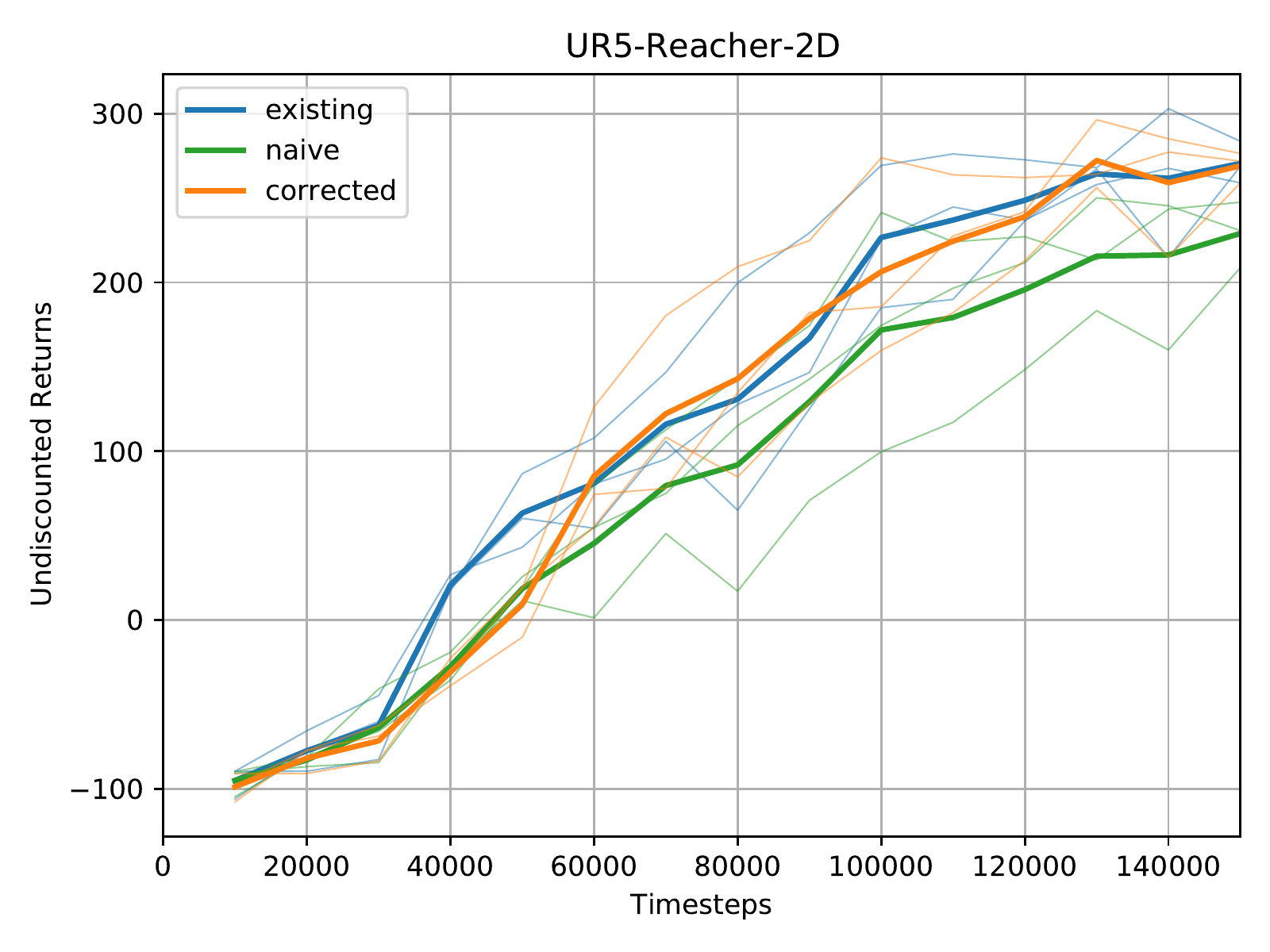}
    \caption{Performance of modified PPO and two baselines on UR5-Reacher task. The solid lines for each variant are averaged over three independent runs.}
    \label{fig:ur5}
    \vspace*{-3mm}
\end{figure*}

We use the Reacher task with UR5 developed by Mahmood et al. 2018 called the UR-Reacher-2 task. 
In the OpenAI Gym Reacher, the agent’s objective is to reach arbitrary target positions by exercising low-level control over a two-joint robotic arm. 
In UR-Reacher-2, we actuate the second and the third joints from the base by sending angular speeds between $[-0.3, +0.3]$ rad/s. 
The observation vector consists of joint angles, joint velocities, the previous action, and the vector difference between the target and the fingertip coordinates. The performance of modified PPO and two baselines on UR-Reacher-2 task is shown in Fig. \ref{fig:ur5}.
The control policy was learned real-time from scratch on a physical robot, where each independent run requires 3h.
We use the same hyper-parameters mentioned earlier for PPO experiments on dm Reacher, except for two small modifications. The weight loss coefficient is set to $1$ and the scaling factor is set to 10. 

\subsection{Analysis on Distribution Distances }

\begin{lemma}
For any policy $\pi$, any positive constant $\epsilon$, and any MDP where the undiscounted state stationary distribution exists, if the total variation between the initial and undiscounted state distributions, $\rho$ and $d_{\pi}$ is small, that is,
$$d_{TV}(\rho,d_{\pi}) \le \epsilon,$$
then the total variation between the discounted and undiscounted state distributions, $d_{\pi}$ and $d_{\pi,\gamma}$, is also small, that is,
$$d_{TV}(d_{\pi,\gamma},d_{\pi}) \le \epsilon.$$
\end{lemma}

\begin{proof}
Recall that the total variation equals
$$d_{TV}(\rho,d_{\pi}) = \frac{1}{2} \sum_s |\rho(s) - d_{\pi}(s)|.$$

First, we can rewrite the undiscounted stationary state distribution as $$d_{\pi} = (1-\gamma)\sum_{t\ge 0 }\gamma^t d_{\pi}(s) \mathbb{P}_{\pi}(S_{t}=s'|S_0=s),$$ 
thanks to its stationary property $\sum_s d_{\pi}(s) \mathbb{P}_{\pi}(S'=s'|S=s) = d_{\pi}(s').$

Next, let us compare the discounted and undiscounted state distributions.
\begin{align*}
    d_{TV}(d_{\pi,\gamma},d_{\pi}) 
    & = \frac{1}{2} \sum_{s'} |\sum_s (1-\gamma)\sum_{t\ge 0 }\gamma^t \rho(s) \mathbb{P}_{\pi}(S_{t}=s'|S_0=s) - \sum_s (1-\gamma)\sum_{t\ge 0 }\gamma^t d_{\pi}(s) \mathbb{P}_{\pi}(S_{t}=s'|S_0=s)|\\
    & =\frac{1}{2} \sum_{s'} (1-\gamma)\sum_{t\ge 0 }\gamma^t |\sum_s (\rho(s)-d_{\pi}(s)) \mathbb{P}_{\pi}(S_t=s'|S_0=s) |\\
    & \le \frac{1}{2} \sum_{s'} (1-\gamma)\sum_{t\ge 0 }\gamma^t  \sum_s   \mathbb{P}_{\pi}(S_t=s'|S_0=s)| \rho(s)-d_{\pi}(s)|\\
    & = \frac{1}{2} (1-\gamma)\sum_{t\ge 0 }\gamma^t \sum_s \sum_{s'}  \mathbb{P}_{\pi}(S_t=s'|S_0=s)| \rho(s)-d_{\pi}(s)|\\
    & = \frac{1}{2} (1-\gamma)\sum_{t\ge 0 }\gamma^t \sum_s | \rho(s)-d_{\pi}(s)|\\
    & = (1-\gamma)\sum_{t\ge 0 }\gamma^t d_{TV}(\rho,d_{\pi})\\
    & \le \epsilon.
\end{align*}
    
\end{proof}

\section{Results Under Discounted Returns}
First, we show the learning curves for our modified batch-actor-critic and corresponding baselines in Figure 14-16. 

The results for three discount factor values are similar. All three algorithms show similar performances on Acrobot. But BAC with our correction in orange dominates CartPole and continuous MountainCar. 

\begin{figure}
    \small
    \centering
    \includegraphics[width=\textwidth]{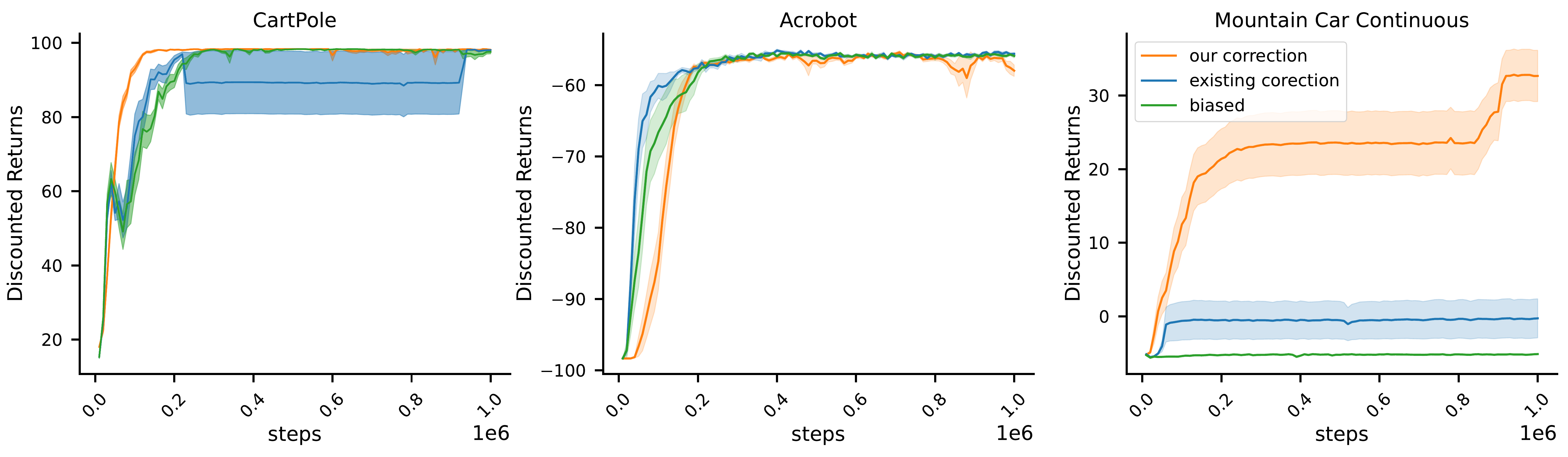}
    \caption{Learning results for BAC with discount factor $0.99$.}
    \includegraphics[width=\textwidth]{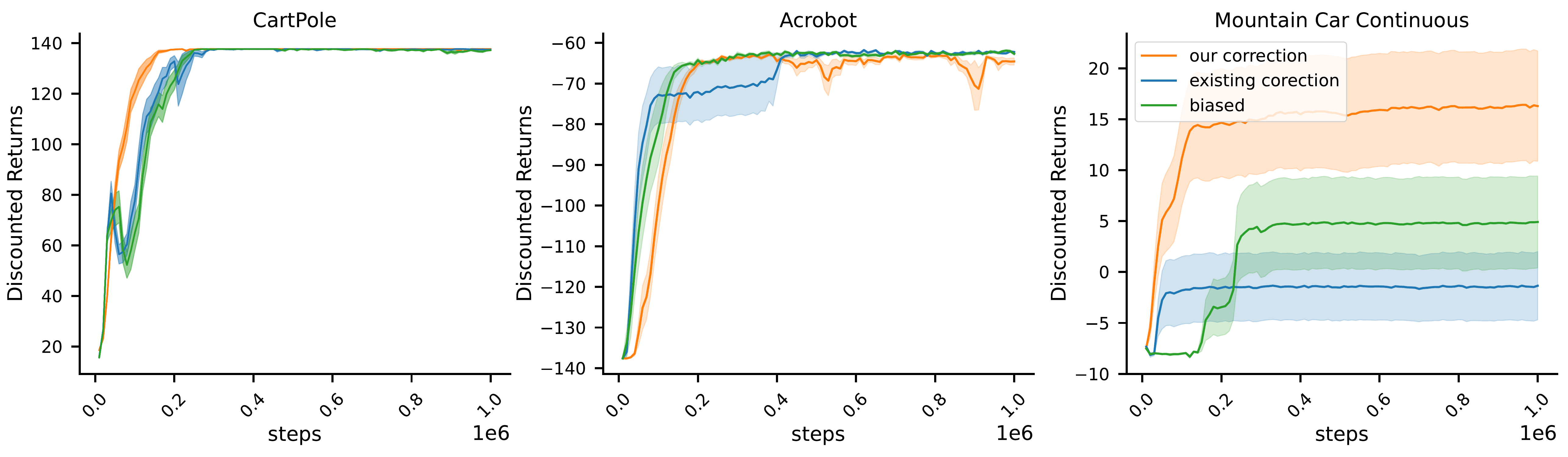}
    \caption{Learning results for BAC with discount factor $0.993$.}
    \includegraphics[width=\textwidth]{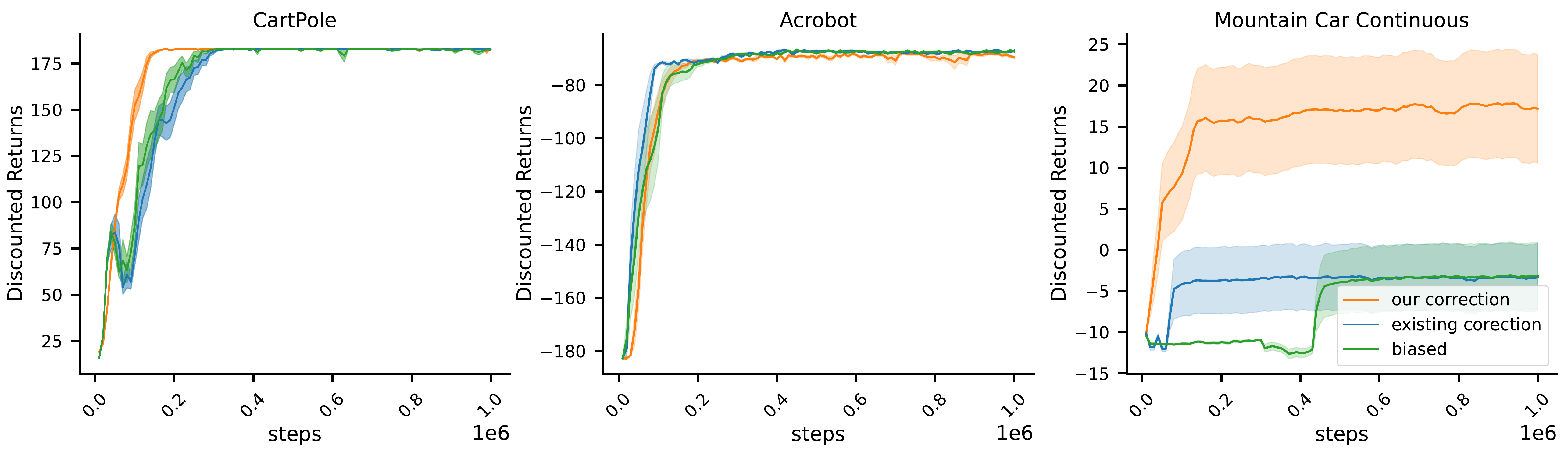}
    \caption{Learning results for BAC with discount factor $0.995$.}
    \label{fig:classic_control}
\end{figure}

\begin{figure}
    \small
    \centering
    \includegraphics[width=\textwidth]{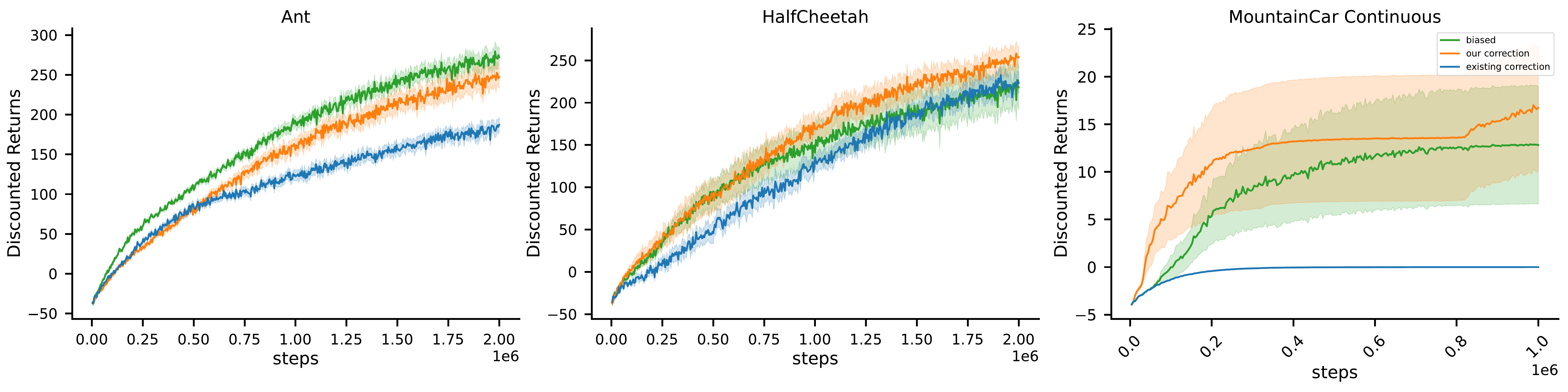}
    \caption{Learning results for PPO with discount factor $0.99$.}
    \includegraphics[width=\textwidth]{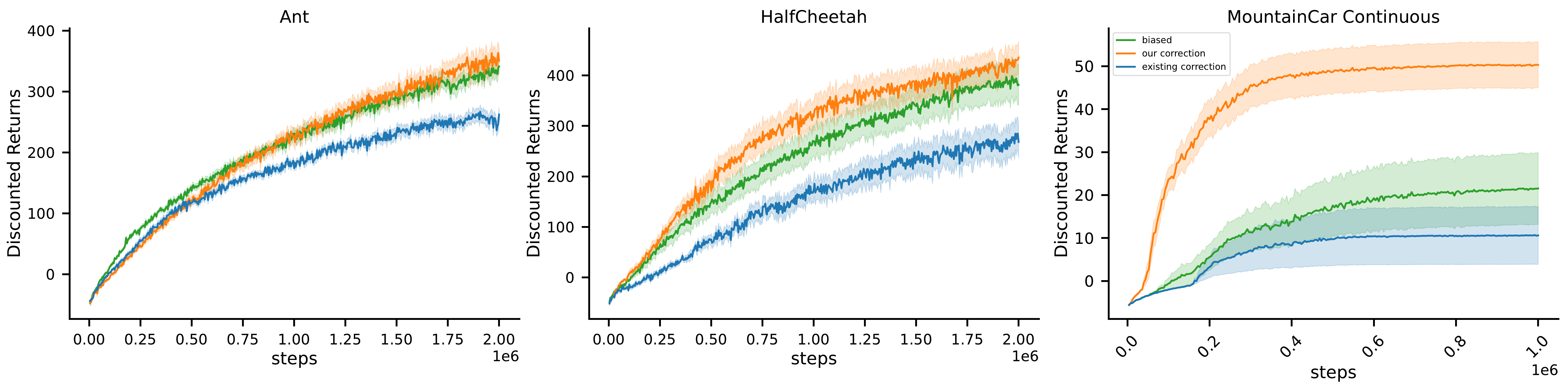}
    \caption{Learning results for PPO with discount factor $0.993$.}
    \includegraphics[width=\textwidth]{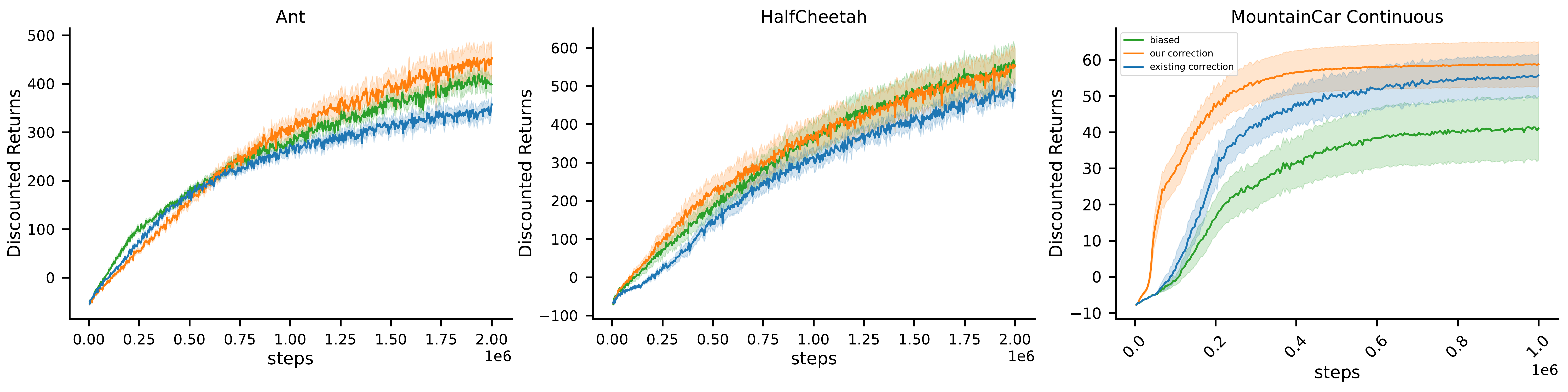}
    \caption{Learning results for PPO with discount factor $0.995$.}
    \label{fig:ppo}
\end{figure}

The learning curves for our modified PPO and corresponding baselines in Figure 17-19. The existing correction hinders the final performance of the agent as shown in blue. It causes worse returns than the original algorithm, except for MountainCar Continuous with discount factor $0.995$. However, PPO with our averaging correction in orange successfully matches the biased algorithm’s performance and can even improves the learning speed and final performance in MountainCar Continuous.




\end{document}